\documentclass[final]{cvpr}

\usepackage{times}
\usepackage{epsfig}
\usepackage{graphicx}
\usepackage{amsmath}
\usepackage{amssymb}
\usepackage{graphicx}
\usepackage{multirow}
\usepackage[title]{appendix}

\usepackage{amsmath,amssymb}

\usepackage{color}
\usepackage{cite}
\usepackage{dsfont}
\usepackage[table,xcdraw]{xcolor}
\usepackage{ragged2e}
\usepackage{nccmath}
\usepackage{textcomp}
\usepackage{arydshln}
\usepackage{amsmath}
\usepackage{caption}

\usepackage[pagebackref=true,breaklinks=true,colorlinks,bookmarks=false]{hyperref}

\def\cvprPaperID{1523} 
\def\confYear{CVPR 2021}
\newcommand{\pX}{\mathbf{x}}
\newcommand{\I}{\mathbf{I}}
\newcommand{\Iy}{\I_y}
\newcommand{\cR}{\textrm{R}}
\newcommand{\cG}{\textrm{G}}
\newcommand{\cB}{\textrm{B}}

\begin{document}

\title{HistoGAN:\\Controlling Colors of GAN-Generated and Real Images via Color Histograms}

\author{\hspace{3mm}Mahmoud Afifi \hspace{10mm} Marcus A. Brubaker \hspace{10mm} Michael S. Brown\vspace{3mm}\\
York University\vspace{2mm}\\
{\tt\small \{mafifi,mab,mbrown\}@eecs.yorku.ca}}

\twocolumn[{%
\renewcommand\twocolumn[1][]{#1}%
\maketitle
\begin{center}
\includegraphics[width=\textwidth]{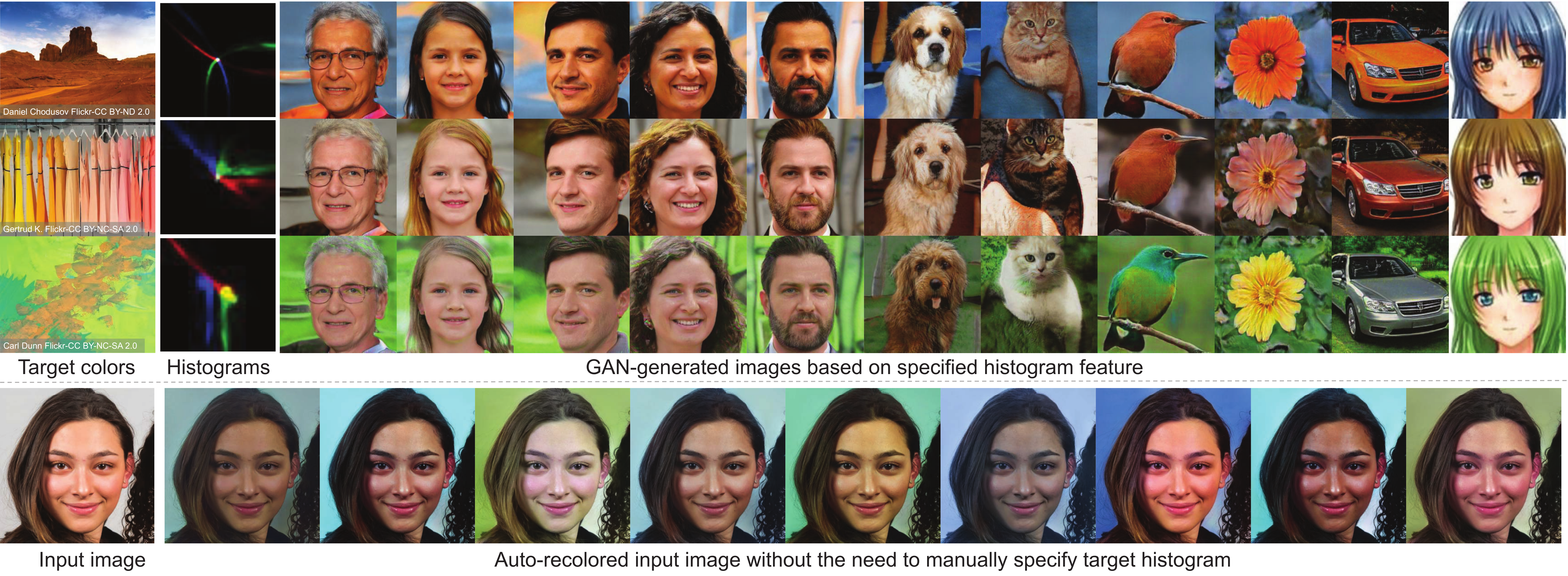}
\vspace{-4.5mm}
\captionof{figure}{HistoGAN is a generative adversarial network (GAN) that learns to manipulate image colors based on histogram features. Top: GAN-generated images with color distributions controlled via target histogram features (left column). Bottom: Results of ReHistoGAN, an extension of HistoGAN to recolor real images, using sampled target histograms.} \label{fig:teaser}
\end{center}%
}]
\maketitle

%
%
\begin{abstract}
While generative adversarial networks (GANs) can successfully produce high-quality images, they can be challenging to control.\  Simplifying GAN-based image generation is critical for their adoption in graphic design and artistic work.\  This goal has led to significant interest in methods that can intuitively control the appearance of images generated by GANs.\ In this paper, we present HistoGAN, a color histogram-based method for controlling GAN-generated images' colors.\ We focus on color histograms as they provide an intuitive way to describe image color while remaining decoupled from domain-specific semantics. Specifically, we introduce an effective modification of the recent StyleGAN architecture~\cite{karras2020analyzing} to control the colors of GAN-generated images specified by a target color histogram feature.\ We then describe how to expand HistoGAN to recolor real images.\  For image recoloring, we jointly train an encoder network along with HistoGAN.\ The recoloring model, ReHistoGAN, is an unsupervised approach trained to encourage the network to keep the original image's content while changing the colors based on the given target histogram. We show that this histogram-based approach offers a better way to control GAN-generated and real images' colors while producing more compelling results compared to existing alternative strategies. 
\end{abstract}


\section{Motivation and Related Work} \label{sec.intro}

Color histograms are an expressive and convenient representation of an image's color content. Color histograms are routinely used by conventional color transfer methods (e.g., \cite{reinhard2001color, xiao2006color, nguyen2014illuminant, faridul2016colour}).
These color transfer methods aim to manipulate the colors in an input image to match those of a target image, such that the images share a similar ``look and feel''. In the color transfer literature, there are various forms of color histograms used to represent the color distribution of an image, such as a direct 3D histogram~\cite{reinhard2001color, xiao2006color,faridul2016colour}, 2D histogram \cite{avi2020deephist, CCC, afifi2019color, afifi2019sensor}, color palette \cite{chang2015palette, zhang2017palette, afifi2019image} or color triad \cite{shugrina2020nonlinear}. Despite the effectiveness of color histograms for color transfer, recent deep learning methods almost exclusively rely on image-based examples to control colors. While image exemplars impact the final colors of generative adversarial network (GAN)-generated images and deep recolored images, these methods -- that mostly target image style transfer -- also affect other style attributes, such as texture information and tonal values~\cite{gatys2015neural, gatys2016image, johnson2016perceptual, ulyanov2016instance, isola2017image, luan2017deep, sheng2018avatar}.
Consequently, the quality of the results produced by these methods often depends on the semantic similarity between the input and target images, or between a target image and a particular domain~\cite{sheng2018avatar, he2019progressive}.

\begin{figure*}[t]
\centering
\includegraphics[width=\linewidth]{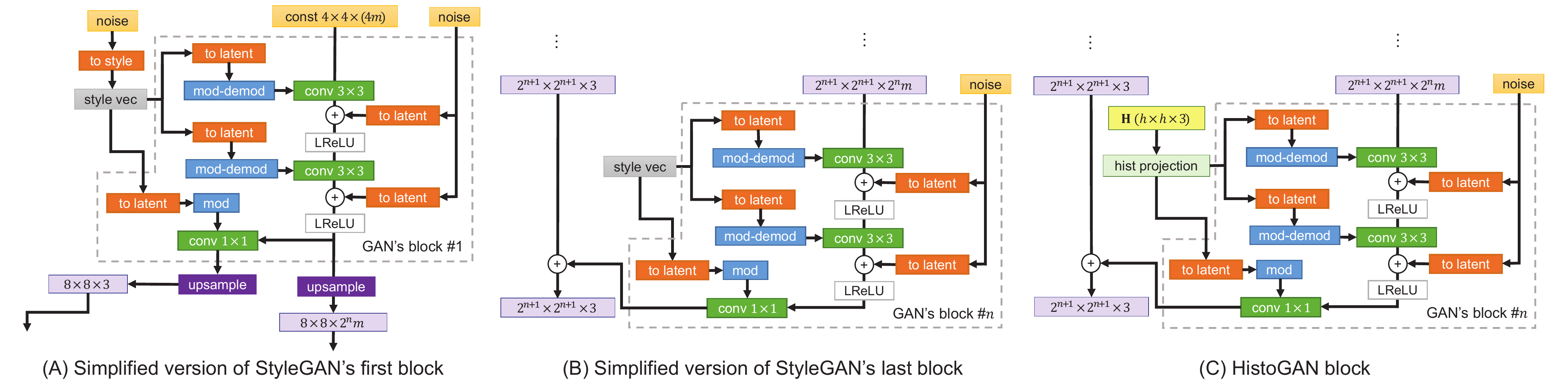}
\vspace{-2mm}
\caption{We inject our histogram into StyleGAN \cite{karras2020analyzing} to control the generated image colors. (A) and (B) are simplified versions of the StyleGAN's first and last blocks. We modified the last two blocks of the StyleGAN by projecting our histogram feature into each block's latent space, as shown in (C). The parameter $m$ controls the capacity of the model.}
\label{fig:GAN-design}	
\end{figure*}

In this paper, our attention is focused explicitly on controlling only the color attributes of images---this can be considered a sub-category of image style transfer. Specifically, our method does not require shared semantic content between the input/GAN-generated images and a target image or guide image. Instead, our method aims to assist the deep network through color histogram information only\footnote{Project page: \href{https://github.com/mahmoudnafifi/HistoGAN}{https://github.com/mahmoudnafifi/HistoGAN}}.  With this motivation, we first explore using color histograms to control the colors of images generated by GANs.

\vspace{-2mm}
\paragraph{Controlling Color in GAN-Generated Images}

GANs are often used as ``black boxes'' that can transform samples from a simple distribution to a meaningful domain distribution without an explicit ability to control the details/style of the generated images \cite{goodfellow2014generative, radford2015unsupervised, karras2017progressive, arjovsky2017wasserstein, liu2019wasserstein}.
Recently, methods have been proposed to control the style of the GAN-generated images.
For example, StyleGAN~\cite{karras2019style, karras2020analyzing} proposed the idea of ``style mixing'', where different latent style vectors are progressively fed to the GAN to control the style and appearance of the output image.
To transfer a specific style in a target image to GAN-generated images, an optimization process can be used to project the target image to the generator network's latent space to generate images that share some properties with the target image~\cite{abdal2019image2stylegan, karras2020analyzing}.
However, this process requires expensive computations to find the latent code of the target image.
Another direction is to jointly train an encoder-generator network to learn this projection \cite{pidhorskyi2020adversarial, li2020mixnmatch, choi2020stargan}.
More recently, methods have advocated different approaches to control the output of GANs, such as using the normalization flow \cite{abdal2020styleflow}, latent-to-domain-specific mapping  \cite{choi2020stargan}, deep classification features \cite{shocher2020semantic}, few-shot image-to-image translation \cite{saito2020coco}, and a single-image training strategy \cite{shaham2019singan}.
Despite the performance improvements, most of these methods are limited to work with a single domain of both target and GAN-generated images \cite{li2020mixnmatch, pidhorskyi2020adversarial}.

We seek to control GAN-generated images using color histograms as our specified representation of image style.
Color histograms enable our method to accept target images taken from \textit{any} arbitrary domain.
Figure \ref{fig:teaser}-top shows GAN-generated examples using our method.
As shown in Fig.~\ref{fig:teaser}, our generated images share the same color distribution as the target images without being restricted to, or influenced by, the semantic content of the target images.

\vspace{-2mm}
\paragraph{Recoloring Real Images}
In addition to controlling the GAN-generated images, we seek to extend our approach to perform image recoloring within the GAN framework. In this context, our method accepts a real input image and a target histogram to produce an output image with the fine details of the input image but with the same color distribution given in the target histogram.
Our method is trained in a fully unsupervised fashion, where no ground-truth recolored image is required.
Instead, we propose a novel adversarial-based loss function to train our network to extract and consider the color information in the given target histogram while producing realistic recolored images.
One of the key advantages of using the color histogram representation as our target colors can be shown in Fig.\ \ref{fig:teaser}-bottom, where we can {\it automatically recolor} an image without directly having to specify a target color histogram.
Auto-image recoloring is a less explored research area with only a few attempts in the literature (e.g., \cite{laffont2014transient, deshpande2017learning, afifi2019image, anokhin2020high}).

\section{HistoGAN} \label{sec.method}

We begin by describing the histogram feature used by our method (Sec.\ \ref{subsec.histoblock}). Afterwards, we discuss the proposed modification to StyleGAN \cite{karras2020analyzing} to incorporate our histogram feature into the generator network (Sec.\ \ref{subsec.method-coloring-GAN-images}). Lastly, we explain how this method can be expanded to control colors of real input images to perform image recoloring~(Sec.\ \ref{subsec.method-recoloring}).

\subsection{Histogram feature} \label{subsec.histoblock}
The histogram feature used by HistoGAN is borrowed from the color constancy literature \cite{CCC, afifi2019color, afifi2019sensor} and is constructed to be a differentiable histogram of colors in the log-chroma space due to better invariance to illumination changes~\cite{finlayson2001color, eibenberger2012importance}. The feature is a 2D histogram of an image's colors projected into a log-chroma space.
This 2D histogram is parameterized by $uv$ and conveys an image's color information while being more compact than a typical 3D histogram defined in RGB space.   A log-chroma space is defined by the intensity of one channel, normalized by the other two, giving three possible options of how it is defined.
Instead of selecting only one such space, all three options can be used to construct three different histograms which are combined together into a histogram feature, $\mathbf{H}$, as an $h\!\times\!h\!\times 3$ tensor~\cite{afifi2019color} .

The histogram is computed from a given input image, $\I$, by first converting it into the log-chroma space.
For instance, selecting the $\cR$ color channel as primary and normalizing by $\cG$ and $\cB$ gives:
\begin{equation}
\resizebox{0.9\hsize}{!}{
$\I_{uR}(\pX) = \log{\left(\frac{ \I_{\cR}(\pX)+ \epsilon}{\I_{\cG}(\pX)+ \epsilon}\right)} \textrm{ , }  \I_{vR}(\pX) = \log{\left(\frac{ \I_{\cR}(\pX)+ \epsilon}{\I_{\cB}(\pX)+ \epsilon}\right)},$}
\end{equation}
where the $\cR, \cG, \cB$ subscripts refer to the color channels of the image $\I$, $\epsilon$ is a small constant added for numerical stability, $\pX$ is the pixel index, and $(uR, vR)$ are the $uv$ coordinates based on using $\cR$ as the primary channel.
The other components $\I_{uG}$, $\I_{vG}$, $\I_{uB}$, $\I_{vB}$ are computed similarly by projecting the $\cG$ and $\cB$ color channels to the log-chroma space.
In \cite{afifi2019color}, the RGB-$uv$ histogram is computed by thresholding colors to a bin and computing the contribution of each pixel based on the intensity $\Iy(\pX) = \sqrt{\I_{\cR}^{2}(\pX) + \I_{\cG}^{2}(\pX) + \I_{\cB}^{2}(\pX)}$.  In order to make the representation differentiable, \cite{afifi2019sensor} replaced the thresholding operator with a kernel weighted contribution to each bin.  The final unnormalized histogram is computed as:
\begin{equation}
\mathbf{H}(u,v,c) \propto \sum_{\pX} k(\I_{uc}(\pX), \I_{vc}(\pX), u, v) \I_{y}(\pX),
\end{equation}
where $c \in {\{\textrm{R}, \textrm{G}, \textrm{B}\}}$ and $k(\cdot)$ is a pre-defined kernel.
While a Gaussian kernel was originally used in \cite{afifi2019sensor}, we found that the inverse-quadratic kernel significantly improved training stability.  The inverse-quadratic kernel is defined as:
\begin{multline}
k(\I_{uc}, \I_{vc}, u, v) = \left(1+\left(\left| \I_{uc} - u \right|/\tau\right)^2\right)^{-1} \\
\times \left(1 + \left(\left| \I_{vc} - v \right|/\tau\right)^2\right)^{-1},
\end{multline}
\noindent
where  $\tau$ is a fall-off parameter to control the smoothness of the histogram's bins.
Finally, the histogram feature is normalized to sum to one, i.e., $\sum_{u,v,c}\mathbf{H}(u,v,c)=1$.

\begin{figure}
\centering
\includegraphics[width=\linewidth]{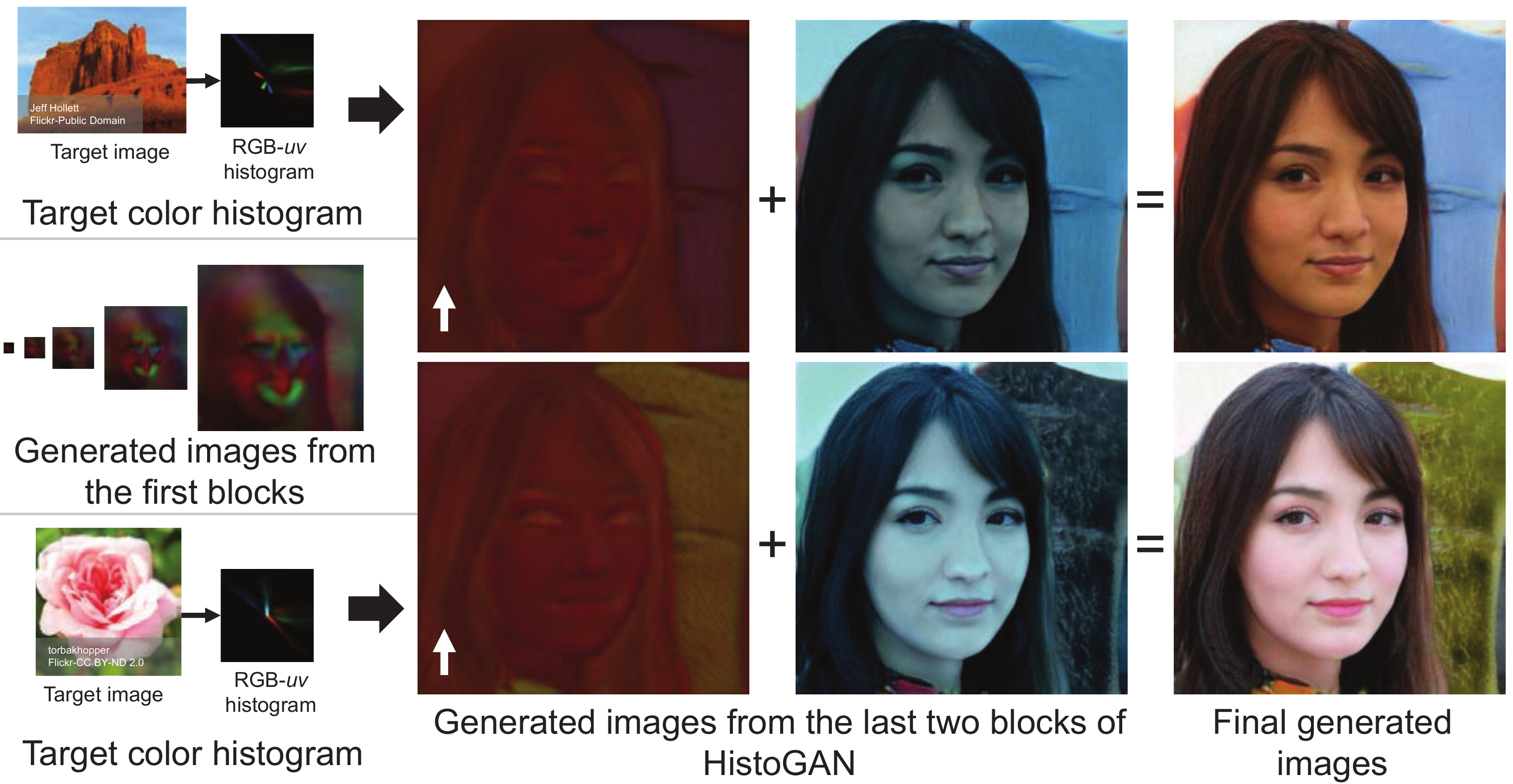}
\vspace{-6.5mm}
\caption{Progressively generated images using the HistoGAN modifications.}
\label{fig:analysis}
\end{figure}

\subsection{Color-controlled Image Generation}
\label{subsec.method-coloring-GAN-images}

Our histogram feature is incorporated into an architecture based on StyelGAN~\cite{karras2020analyzing}.
Specifically, we modified the original design of StyleGAN (Fig.\ \ref{fig:GAN-design}-[A] and [B]) such that we can ``inject'' the histogram feature into the progressive construction of the output image.
The last two blocks of the StyleGAN (Fig.\ \ref{fig:GAN-design}-[B]) are modified by replacing the fine-style vector with the color histogram feature.
The histogram feature is then projected into a lower-dimensional representation by a ``histogram projection'' network (Fig.\ \ref{fig:GAN-design}-[C]).
This network consists of eight fully connected layers with a leaky ReLU (LReLU) activation function \cite{maas2013rectifier}.
The first layer has 1,024 units, while each of the remaining seven layers has 512.
The ``to-latent'' block, shown in orange in Fig.\ \ref{fig:GAN-design}, maps the projected histogram to the latent space of each block.  This ``to-latent'' block consists of a single fc layer with $2^n m$ output neurons, where $n$ is the block number, and $m$ is a parameter used to control the entire capacity of the network.

To encourage generated images to match the target color histogram, a color matching loss is introduced to train the generator.
Because of the differentiability of our histogram representation, the loss function, $C(\mathbf{H}_g,\mathbf{H}_t)$, can be any differentiable metric of similarity between the generated and target histograms $\mathbf{H}_g$ and $\mathbf{H}_t$, respectively.
For simplicity, we use the Hellinger distance defined as:
\begin{equation}\label{eq:hellinger-distance}
C\left(\mathbf{H}_g, \mathbf{H}_t\right) = \frac{1}{\sqrt{2}} \left\Vert \mathbf{H}_g^{1/2} - \mathbf{H}_t^{1/2} \right\Vert_2,
\end{equation}
where $\Vert \cdot \Vert_2$ is the standard Euclidean norm and $\mathbf{H}^{1/2}$ is an element-wise square root. Note that the Hellinger distance is closely related to the Bhattacharyya coefficient, $B(\cdot)$, where $C\left(\mathbf{H}_g, \mathbf{H}_t\right) = \left(1-B\left(\mathbf{H}_g, \mathbf{H}_t\right)\right)^{1/2}$.

This color-matching histogram loss function is combined with the discriminator to give the generator network loss:
\begin{equation}\label{eq:gan-loss}
{\mathcal{L}}_g = D\left(\I_g\right) + \alpha C\left(\mathbf{H}_g, \mathbf{H}_t\right),
\end{equation}
where $\I_g$ is the GAN-generated image, $D\left(\cdot\right)$ is our discriminator network that produces a scalar feature given an image (see supp.\ materials for more details), $\mathbf{H}_t$ is the target histogram feature (injected into the generator network), $\mathbf{H}_g$ is the histogram feature of $\I_g$, $C\left(\cdot\right)$ is our histogram loss function, and $\alpha$ is a scale factor to control the strength of the histogram loss term.

\begin{figure}
\centering
\includegraphics[width=\linewidth]{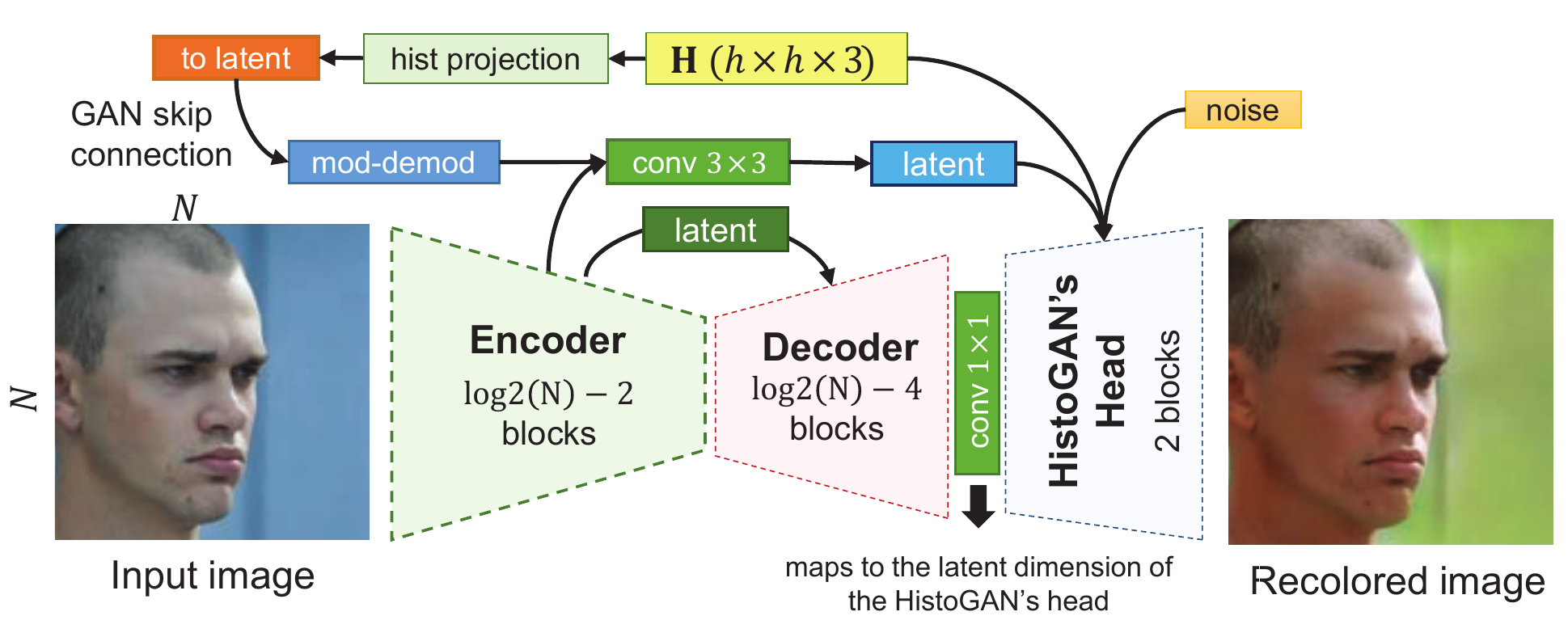}
\vspace{-6.5mm}
\caption{Our Recoloring-HistoGAN (ReHistoGAN) network. We map the input image into the HistoGAN's latent space using an encoder-decoder network with skip connections between each encoder and decoder blocks. Additionally, we pass the latent feature of the first two encoder blocks to our GAN's head after processing it with the histogram's latent feature.}\vspace{-2mm}
\label{fig:recoloring-design}
\end{figure}

As our histogram feature is computed by a set of differentiable operations, our loss function (Eqs. \ref{eq:hellinger-distance} and \ref{eq:gan-loss}) can be optimized using SGD.
During training, different target histograms $\mathbf{H}_t$ are required. To generate these for each generated image, we randomly select two images from the training set, compute their histograms $\mathbf{H}_1$ and $\mathbf{H}_2$, and then randomly interpolate between them. Specifically, for each generated image during training, we generate a random target histogram as follows:
\begin{equation}
\label{eq.target_hist}
\mathbf{H}_t = \delta \mathbf{H}_1 + \left(1 - \delta \right) \mathbf{H}_2,
\end{equation}
where $\delta \sim U(0,1)$ is sampled uniformly. The motivation behind this interpolation process is to expand the variety of histograms during training. This is a form of data augmentation for the histograms with the implicit assumption of the convexity of the histogram distribution in the target domain (e.g., face images). We found this augmentation helped reduce overfitting to the histograms of the training images and ensured robustness at test time. We note that this assumption does not hold true for target domains with high diversity where the target histograms span a broad range in the log-chroma space and can be multimodal (e.g., landscape images). Nonetheless, we found that even in those cases the augmentation was still beneficial to the training.

With this modification to the original StyleGAN architecture, our method can control the colors of generated images using our color histogram features.\ Figure~\ref{fig:analysis} shows the progressive construction of the generated image by HistoGAN.  As can be seen, the outputs of the last two blocks are adjusted to consider the information conveyed by the target histogram to produce output images with the same color distribution represented in the target histogram.

\begin{figure}
\includegraphics[width=\linewidth]{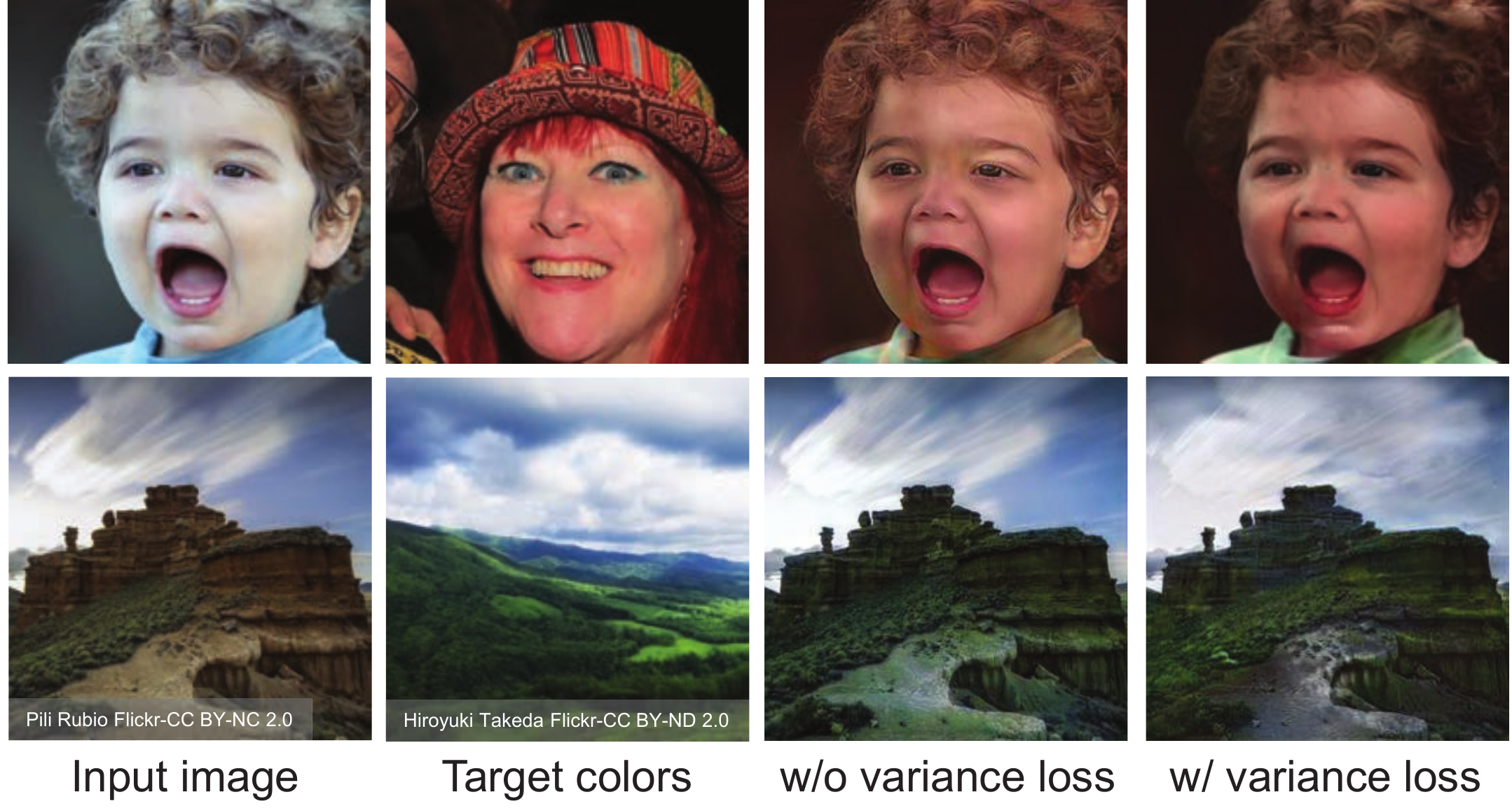}
\vspace{-6.5mm}
\caption{Results of training ReHistoGAN with and without the variance loss term described in Eq. \ref{eq.variance-loss}.}\vspace{-2mm}
\label{fig:variance_loss}
\end{figure}

\begin{figure}[b]
\centering
\includegraphics[width=\linewidth]{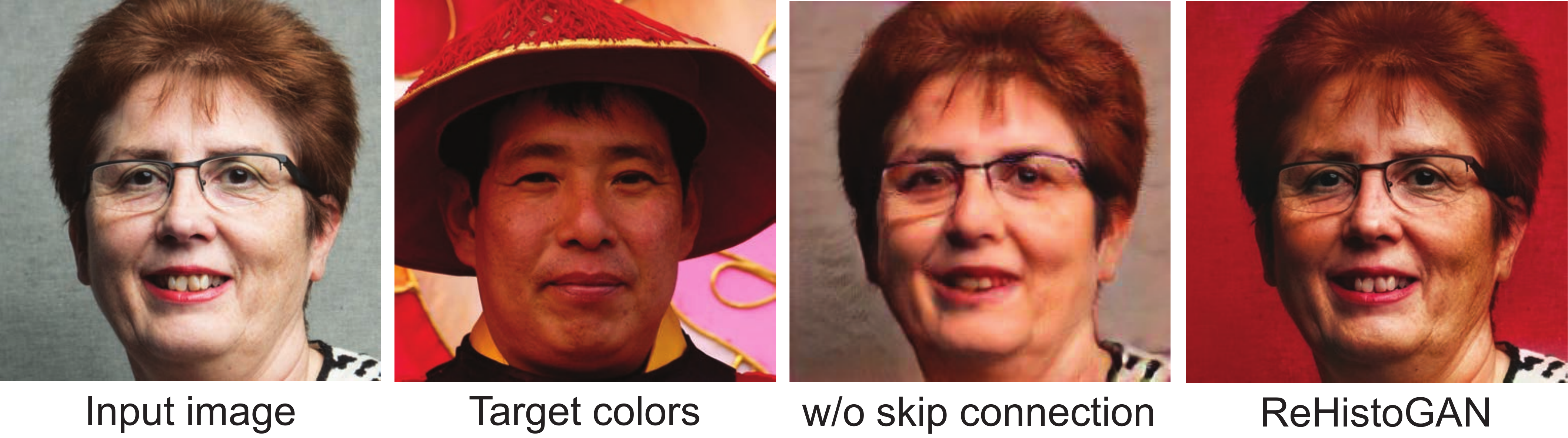}
\vspace{-6.5mm}
\caption{Results of image recoloring using the encoder-GAN reconstruction without skip connections and our ReHistoGAN using our proposed loss function.}\vspace{-2mm}
\label{fig:comparison_with_projection}
\end{figure}

\begin{figure*}
\centering
\includegraphics[width=\linewidth]{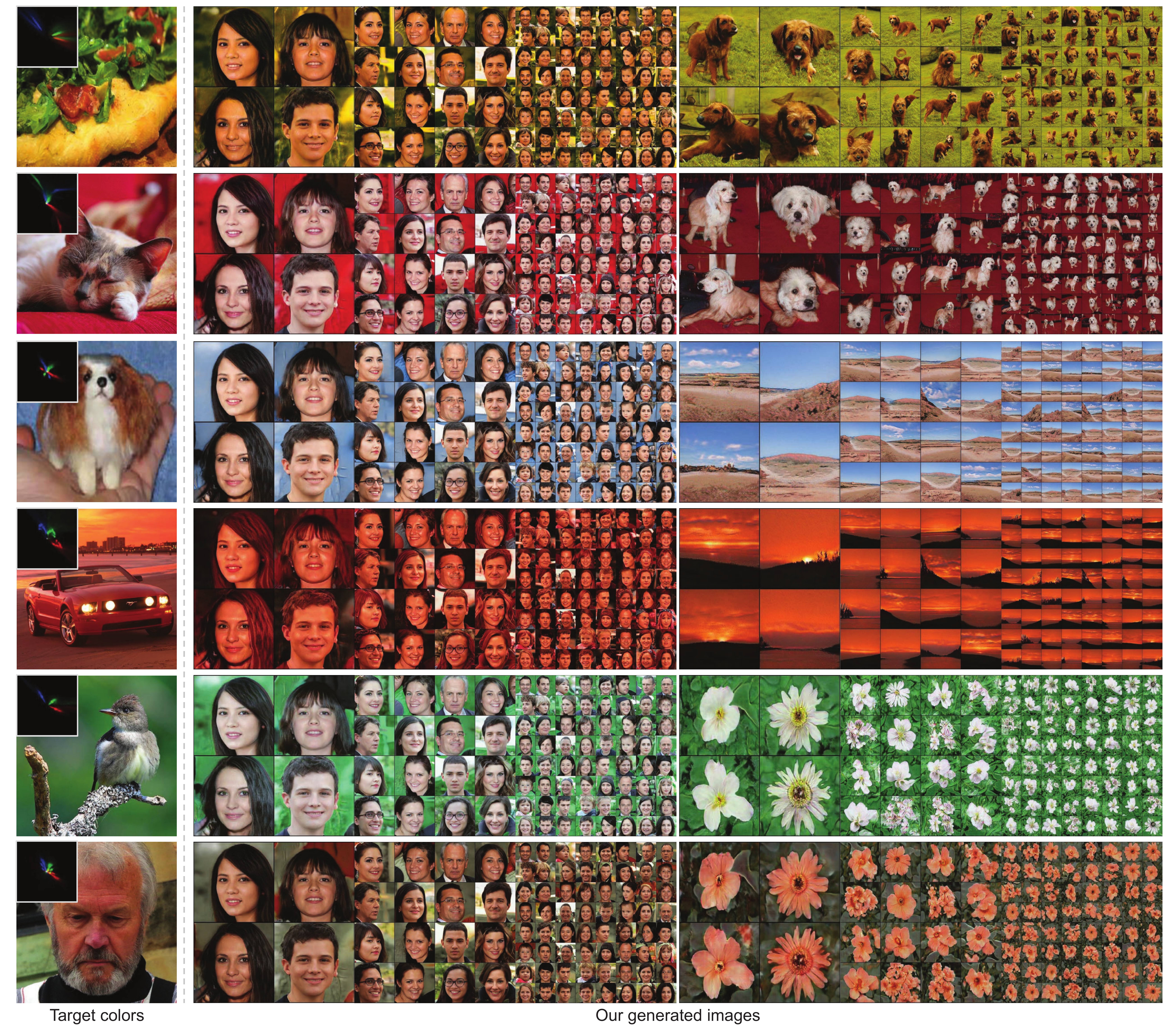}
\vspace{-6.5mm}
\caption{Images generated by HistoGAN. For each input image shown in the left, we computed the corresponding target histogram (shown in the upper left corner of the left column) and used it to control colors of the generated images in each row.}\vspace{-2mm}
\label{fig:GAN_results}
\end{figure*}

\subsection{Image Recoloring} \label{subsec.method-recoloring}

We can also extend HistoGAN to recolor an input image, as shown in Fig.\ \ref{fig:teaser}-bottom. Recoloring an existing input image, $\I_i$, is not straightforward because the randomly sampled noise and style vectors are not available as they are in a GAN-generated scenario.  As shown in Fig.\ \ref{fig:analysis}, the head of HistoGAN (i.e., the last two blocks) are responsible for controlling the colors of the output image.
Instead of optimizing for noise and style vectors that could be used to generate a given image $\I_i$, we propose to train an encoding network that maps the input image into the necessary inputs of the head of HistoGAN.
With this approach, the head block can be given different histogram inputs to produce a wide variety of recolored versions of the input image.
We dub this extension the ``Recoloring-HistoGAN'' or ReHistoGAN for short.
The architecture of ReHistoGAN is shown in Fig.\ \ref{fig:recoloring-design}.
The ``encoder'' has a U-Net-like structure \cite{ronneberger2015u} with skip connections.
To ensure that fine details are preserved in the recolored image, $\I_r$, the early latent feature produced by the first two U-Net blocks are further provided as input into the HistoGAN's head through skip connections.


The target color information is passed to the HistoGAN head blocks as described in Sec.\ \ref{subsec.method-coloring-GAN-images}. Additionally, we allow the target color information to influence through the skip connections to go from the first two U-Net-encoder blocks to the HistoGAN's head.
We add an additional histogram projection network, along with a ``to-latent'' block, to project our target histogram to a latent representation.
This latent code of the histogram is processed by weight modulation-demodulation operations \cite{karras2020analyzing} and is then convolved over the skipped latent of the U-Net-encoder's first two blocks.
We modified the HistoGAN block, described in Fig.\ \ref{fig:GAN-design}, to accept this passed information (see supp.\ materials for more information).
The leakage of the target color information helps ReHistoGAN to consider information from both the input image content and the target histogram in the recoloring process.

We initialize our encoder-decoder network using He's initialization \cite{he2015delving}, while the weights of the HistoGAN head are initialized based on a previously trained HistoGAN model (trained in Sec.\ \ref{subsec.method-coloring-GAN-images}).
The entire ReHistoGAN is then jointly trained to minimize the following loss function:
\begin{equation}
\label{eq.recoloring-loss}
{\mathcal{L}}_r = \beta R\left(\I_i, \I_r\right) + \gamma D\left(\I_r\right) + \alpha C\left(\mathbf{H}_r, \mathbf{H}_t\right)
\end{equation}
where $R\left(\cdot\right)$ is a reconstruction term, which encourages the preservation of image structure and $\alpha$, $\beta$, and $\gamma$ are hyperparameters used to control the strength of each loss term (see supp.\ materials for associated ablation study).
The reconstruction loss term, $R\left(\cdot\right)$, computes the L1 norm between the second order derivative of our input and recolored images as:
\begin{equation}
	R\left(\I_i, \I_r\right) = \left\Vert \I_i \ast \mathbf{L} - \I_r \ast \mathbf{L} \right\Vert_1
\end{equation}
where $\ast \mathbf{L}$ denotes the application of the Laplacian operator.
The idea of employing the image derivative was used initially to achieve image seamless cloning \cite{perez2003poisson}, where this Laplacian operator suppressed image color information while keeping the most significant perceptual details.
Intuitively, ReHistoGAN is trained to consider the following aspects in the output image: (i) having a similar color distribution to the one represented in the target histogram, this is considered by $C\left(\cdot\right)$, (ii) being realistic, which is the goal of  $D\left(\cdot\right)$, and (iii) having the same content of the input image, which is the goal of $R\left(\cdot\right)$.

Our model trained using the loss function described in Eq.\ \ref{eq.recoloring-loss} produces reasonable recoloring results.
However, we noticed that, in some cases, our model tends to only apply a global color cast (i.e., shifting the recolored image's histogram) to minimize $C\left(\cdot\right)$.
To mitigate this behavior, we added variance loss term to Eq.\ \ref{eq.recoloring-loss}.
The variance loss can be described as:
\begin{equation}
\label{eq.variance-loss}
V(\I_i, \I_r) = - w \sum_{c\in\{\cR,\cG,\cB\}}{\left| \sigma\left(\I_{ic} \ast \mathbf{G}\right) - \sigma\left(\I_{rc} \ast \mathbf{G}\right) \right|},
\end{equation}
where $\sigma\left(\cdot\right)$ computes the standard deviation of its input (in this case the blurred versions of $\I_i$ and $\I_r$ using a Gaussian blur kernel, $\mathbf{G}$, with a scale parameter of $15$), and $w = \Vert \mathbf{H}_t - \mathbf{H}_i \Vert_1$ is a weighting factor that increases as the target histogram and the input image's histogram, $\mathbf{H}_t$ and $\mathbf{H}_i$, become dissimilar and the global shift solution becomes more problematic.
The variance loss encourages the network to avoid the global shifting solution by increasing the differences between the color variance in the input and recolored images. The reason behind using a blurred version of each image is to avoid having a contradiction between the variance loss and the reconstruction loss---the former aims to increase the differences between the variance of the \textit{smoothed} colors in each image, while the latter aims to retain the similarity between the fine details of the input and recolored images. Figure \ref{fig:variance_loss} shows recoloring results of our trained models with and without the variance loss term.

We train ReHistoGAN with target histograms sampled from the target domain dataset, as described earlier in Sec.\ \ref{subsec.method-coloring-GAN-images} (Eq.\ \ref{eq.target_hist}).

A simpler architecture was experimented initially, which did not make use of the skip connections and the end-to-end fine tuning (i.e., the weights of the HistoGAN head were fixed).
However, this approach gave unsatisfactory result, and generally failed to retain fine details of the input image.
A comparison between this approach and the above ReHistoGAN architecture can be seen in Fig.\ \ref{fig:comparison_with_projection}.

\begin{figure}
\centering
\includegraphics[width=\linewidth]{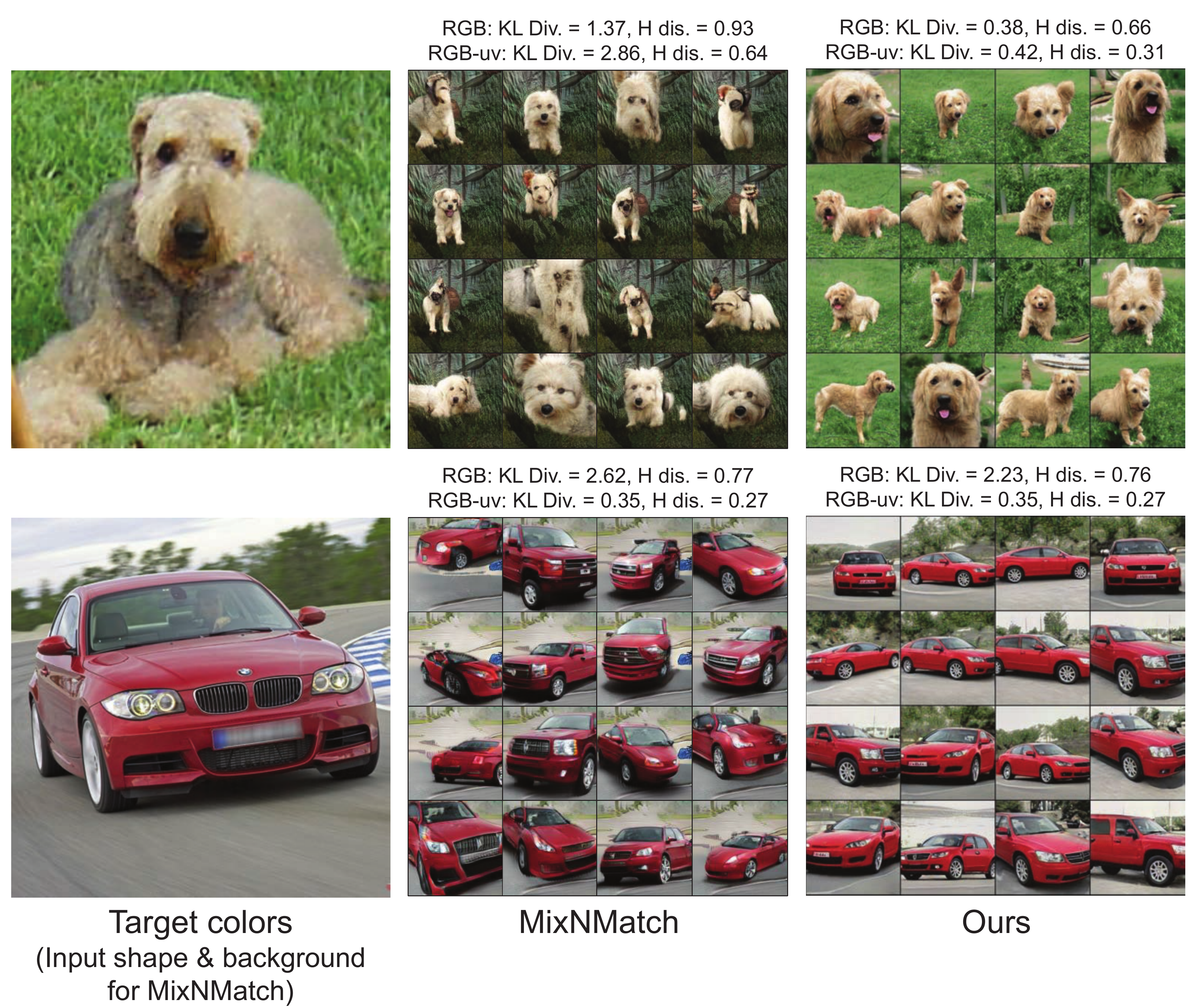}
\vspace{-6.5mm}
\caption{Comparison with the MixNMatch method \cite{li2020mixnmatch}. In the shown results, the target images are used as input shape and background images for the MixNMatch method \cite{li2020mixnmatch}.}\vspace{-2mm}
\label{fig:GAN_comparison_w_MixNMatch}
\end{figure}

\begin{figure*}
\centering
\includegraphics[width=\linewidth]{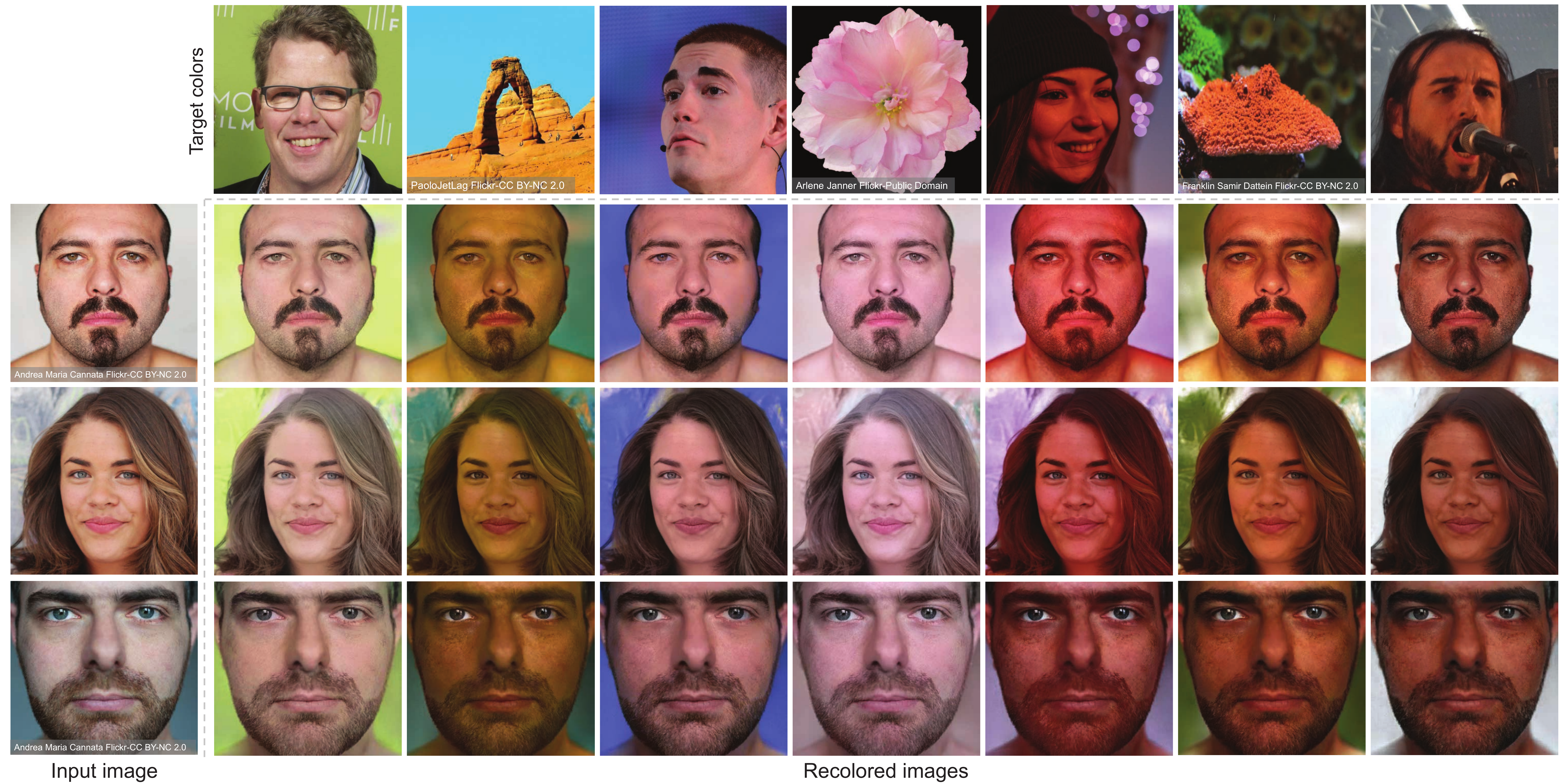}
\vspace{-6.5mm}
\caption{Results of our ReHistoGAN. The shown results are after recoloring input images (shown in the left column) using the target colors (shown in the top row).}\vspace{-1mm}
\label{fig:face_recoloring}
\end{figure*}

\section{Results and Discussion} \label{sec.results}

This section discusses our results and comparisons with alternative methods proposed in the literature for controlling color.
Due to hardware limitations, we used a lightweight version of the original StyleGAN \cite{karras2020analyzing} by setting $m$ to 16, shown in Fig.\ \ref{fig:GAN-design}.
We begin by presenting our image generation results, followed by our results on image recoloring.
Additional results, comparisons, and discussion are also available in the supp.\ materials.

\begin{table*}
\centering
\caption{Comparison with StyleGAN \cite{karras2020analyzing}. The term `w/ proj.' refers to projecting the target image colors into the latent space of StyleGAN. We computed the similarity between the target and generated histograms in RGB and projected RGB-$uv$ color spaces. For each dataset, we report the number of training images. Note that StyleGAN results shown here \textit{do not} represent the actual output of \cite{karras2020analyzing}, as the used model here has less capacity ($m=16$).\label{table:results}}
\vspace{0.5mm}
\scalebox{0.75}{
\begin{tabular}{|c|c|c|c|c|c|c|c|c|c|c|c|}
\hline
\multirow{3}{*}{Dataset} & \multicolumn{6}{c|}{StyleGAN \cite{karras2020analyzing}} & \multicolumn{5}{c|}{HistoGAN (ours)} \\ \cline{2-12}
 & \multicolumn{2}{c|}{FID} & \multicolumn{2}{c|}{RGB hist. (w/ proj.)} & \multicolumn{2}{c|}{RGB-$uv$ hist. (w/ proj.)} & \multirow{2}{*}{FID} & \multicolumn{2}{c|}{RGB hist. (w/ proj.)} & \multicolumn{2}{c|}{RGB-$uv$ hist. (w/ proj.)} \\ \cline{2-7} \cline{9-12}
 & w/o proj. & w/ proj. & KL Div. & H dis. & KL Div. & H dis. &  & KL Div. & H dis. & KL Div. & H dis. \\ \hline
Faces (69,822) \cite{karras2019style} & 9.5018 & 14.194 & 1.3124 & 0.9710 & 1.2125 & 0.6724 & \cellcolor[HTML]{FFFFC7}{\textbf{8.9387}} & \cellcolor[HTML]{FFFFC7}{\textbf{0.9810}} & \cellcolor[HTML]{FFFFC7}{\textbf{0.7487}} & \cellcolor[HTML]{FFFFC7}{\textbf{0.4470}} & \cellcolor[HTML]{FFFFC7}{\textbf{0.3088}} \\ \hline
Flowers (8,189) \cite{nilsback2008automated} & 10.876 & 15.502 & 1.0304 & 0.9614 & 2.7110 & 0.7038 & \cellcolor[HTML]{FFFFC7}{\textbf{4.9572}} & \cellcolor[HTML]{FFFFC7}{\textbf{0.8986}} & \cellcolor[HTML]{FFFFC7}{\textbf{0.7353}} & \cellcolor[HTML]{FFFFC7}{\textbf{0.3837}} & \cellcolor[HTML]{FFFFC7}{\textbf{0.2957}} \\ \hline
Cats (9,992) \cite{catdataset} & \cellcolor[HTML]{FFFFC7}{\textbf{14.366}} & 21.826 & 1.6659 & 0.9740 & 1.4051 & 0.5303 & 17.068 & \cellcolor[HTML]{FFFFC7}{\textbf{1.0054}} & \cellcolor[HTML]{FFFFC7}{\textbf{0.7278}} & \cellcolor[HTML]{FFFFC7}{\textbf{0.3461}} & \cellcolor[HTML]{FFFFC7}{\textbf{0.2639}}\\ \hline
Dogs (20,579) \cite{khosla2011novel} & \cellcolor[HTML]{FFFFC7}{\textbf{16.706}} & 30.403 & 1.9042 & 0.9703 & 1.4856 & 0.5658 & 20.336 & \cellcolor[HTML]{FFFFC7}{\textbf{1.3565}} & \cellcolor[HTML]{FFFFC7}{\textbf{0.7405}} & \cellcolor[HTML]{FFFFC7}{\textbf{0.4321}} & \cellcolor[HTML]{FFFFC7}{\textbf{0.3058}} \\ \hline
Birds (9,053) \cite{wah2011caltech} & 3.5539 & 12.564 & 1.9035 & 0.9706 & 1.9134 & 0.6091 & \cellcolor[HTML]{FFFFC7}{\textbf{3.2251}} & \cellcolor[HTML]{FFFFC7}{\textbf{1.4976}} & \cellcolor[HTML]{FFFFC7}{\textbf{0.7819}} & \cellcolor[HTML]{FFFFC7}{\textbf{0.4261}} & \cellcolor[HTML]{FFFFC7}{\textbf{0.3064}} \\ \hline
Anime (63,565) \cite{animedataset}& \cellcolor[HTML]{FFFFC7}{\textbf{2.5002}} & 9.8890 & 0.9747 & 0.9869 & 1.4323 & 0.5929 & 5.3757 & \cellcolor[HTML]{FFFFC7}{\textbf{0.8547}} & \cellcolor[HTML]{FFFFC7}{\textbf{0.6211}} & \cellcolor[HTML]{FFFFC7}{\textbf{0.1352}} & \cellcolor[HTML]{FFFFC7}{\textbf{0.1798}} \\ \hline
Hands (11,076) \cite{afifi201911k} & 2.6853
 & 2.7826 & 0.9387 & 0.9942 & 0.3654 & 0.3709 & \cellcolor[HTML]{FFFFC7}{\textbf{2.2438}}
 & \cellcolor[HTML]{FFFFC7}{\textbf{0.3317}}
 & \cellcolor[HTML]{FFFFC7}{\textbf{0.3655}}
 & \cellcolor[HTML]{FFFFC7}{\textbf{0.0533}}
 & \cellcolor[HTML]{FFFFC7}{\textbf{0.1085}} \\ \hline
Landscape (4,316) & 24.216 & 29.248 & 0.8811 & 0.9741 & 1.9492 & 0.6265 & \cellcolor[HTML]{FFFFC7}{\textbf{23.549}} & \cellcolor[HTML]{FFFFC7}{\textbf{0.8315}} & \cellcolor[HTML]{FFFFC7}{\textbf{0.8169}} & \cellcolor[HTML]{FFFFC7}{\textbf{0.5445}} & \cellcolor[HTML]{FFFFC7}{\textbf{0.3346}} \\ \hline
Bedrooms (303,116) \cite{yu2015lsun} & 10.599 & 14.673 & 1.5709 & 0.9703 & 1.2690 & 0.5363 & \cellcolor[HTML]{FFFFC7}{\textbf{4.5320}} & \cellcolor[HTML]{FFFFC7}{\textbf{1.3774}} & \cellcolor[HTML]{FFFFC7}{\textbf{0.7278}} & \cellcolor[HTML]{FFFFC7}{\textbf{0.2547}} & \cellcolor[HTML]{FFFFC7}{\textbf{0.2464}} \\ \hline
Cars (16,185) \cite{krause20133d}& 21.485 & 25.496 & 1.6871 & 0.9749 & 0.7364 & 0.4231
& \cellcolor[HTML]{FFFFC7}{\textbf{14.408}}
& \cellcolor[HTML]{FFFFC7}{\textbf{1.0743}}
& \cellcolor[HTML]{FFFFC7}{\textbf{0.7028}}
& \cellcolor[HTML]{FFFFC7}{\textbf{0.2923}}
& \cellcolor[HTML]{FFFFC7}{\textbf{0.2431}}\\ \hline
Aerial Scenes (36,000) \cite{maggiori2017can} & \cellcolor[HTML]{FFFFC7}{\textbf{11.413}} & 14.498 & 2.1142 & 0.9798 & 1.1462 & 0.5158 & 12.602 & \cellcolor[HTML]{FFFFC7}{\textbf{0.9889}} & \cellcolor[HTML]{FFFFC7}{\textbf{0.5887}} & \cellcolor[HTML]{FFFFC7}{\textbf{0.1757}} & \cellcolor[HTML]{FFFFC7}{\textbf{0.1890}}\\ \hline
\end{tabular}}
\end{table*}

\vspace{-2mm}
\paragraph{Image Generation} \label{sec.results-generated-images}

Figure\ \ref{fig:GAN_results} shows examples of our HistoGAN-generated images.\
Each row shows samples generated from different domains using the corresponding input target colors.\
For each domain, we fixed the style vectors responsible for the coarse and middle styles to show our HistoGAN's response to changes in the target histograms.
Qualitative comparisons with the recent MixNMatch method \cite{li2020mixnmatch} are provided in Fig.\ \ref{fig:GAN_comparison_w_MixNMatch}.

To evaluate the potential improvement/degradation of the generated-image diversity and quality caused by our modification to StyleGAN, we trained StyleGAN~\cite{karras2020analyzing} with $m=16$ (i.e., the same as our model capacity) without our histogram modification.
We evaluated both models on different datasets, including our collected set of landscape images.
For each dataset, we generated 10,000 $256\!\times\!256$ images using the StyleGAN and our HistoGAN.
We evaluated the generated-image quality and diversity using the Frech\'et inception distance (FID) metric \cite{heusel2017gans} using the second max-pooling features of the Inception model~\cite{szegedy2015going}.

We further evaluated the ability of StyleGAN to control colors of GAN-generated images by training a regression deep neural network (ResNet \cite{he2016deep}) to transform generated images back to the corresponding fine-style vectors.
These fine-style vectors are used by the last two blocks of StyleGAN and are responsible for controlling delicate styles, such as colors and lights \cite{karras2019style, karras2020analyzing}.

The training was performed for each domain separately using 100,000 training StyleGAN-generated images and their corresponding ``ground-truth'' fine-style vectors.
In the testing phase, we used the trained ResNet to predict the corresponding fine-style vectors of the target image---these target images were used to generate the target color histograms for HistoGAN's experiments. We then generated output images based on the predicted fine-style vectors of each target image.
In the evaluation of StyleGAN and HistoGAN, we used randomly selected target images from the same domain.

The Hellinger distance and KL divergence were used to measure the color errors between the histograms of the generated images and the target histogram; see Table \ref{table:results}.

\begin{figure}[b]
\centering
\vspace{-2mm}
\includegraphics[width=\linewidth]{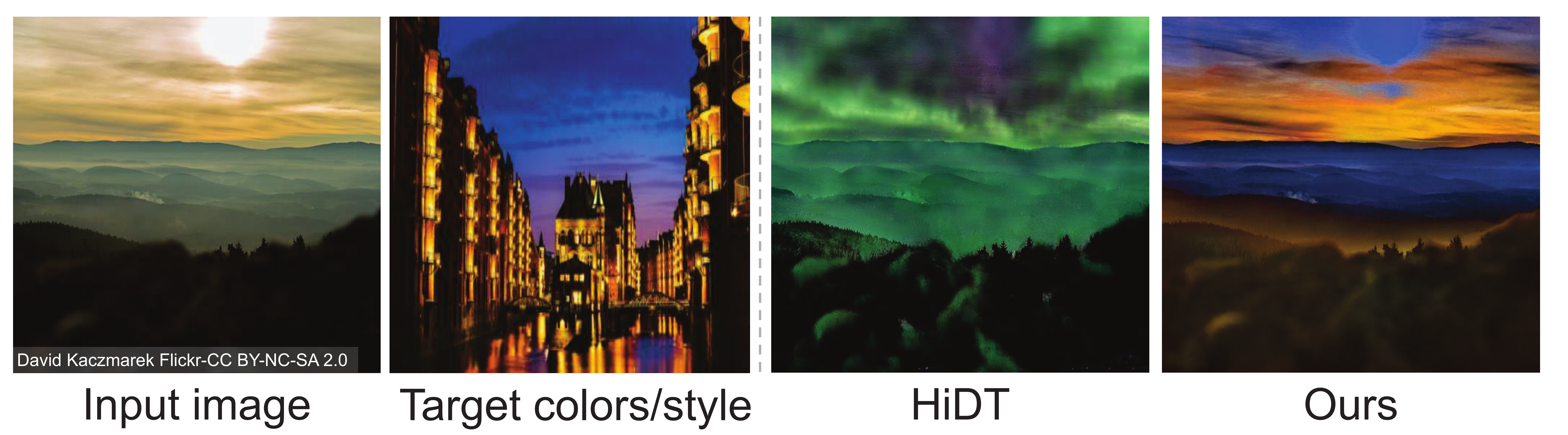}
\vspace{-6.5mm}
\caption{Comparison with the high-resolution daytime translation (HiDT) method \cite{anokhin2020high}.}\vspace{-2mm}
\label{fig:comparison_w_HiDT}
\end{figure}

\begin{figure*}
\centering
\includegraphics[width=\linewidth]{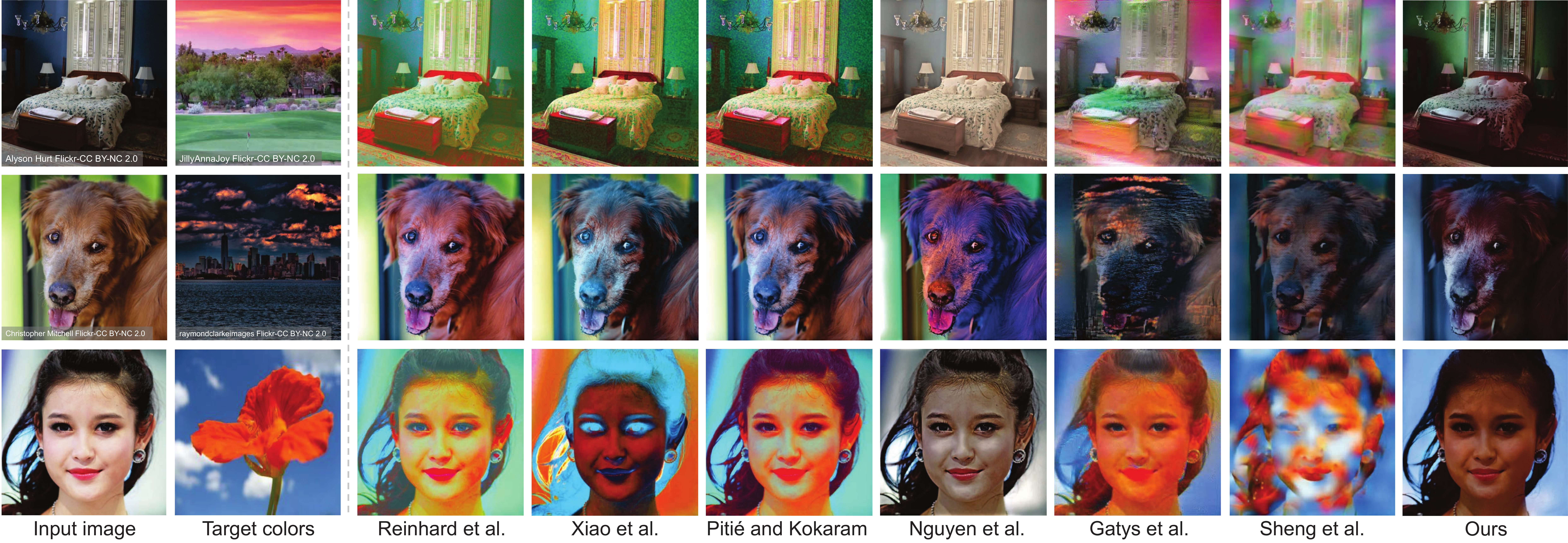}
\vspace{-6mm}
\caption{Comparisons between our ReHistoGAN and other image color/style transfer methods, which are: Reinhard et al., \cite{reinhard2001color}, Xiao et al., \cite{xiao2006color}, Piti\'e and Kokaram \cite{pitie2007}, Nguyen et al., \cite{nguyen2014illuminant}, Gatys et al., \cite{gatys2016image}, and Sheng et al., \cite{sheng2018avatar}.\label{fig:compariosns_recoloring_class_based}}\vspace{-2mm}
\end{figure*}

\vspace{-2mm}
\paragraph{Image Recoloring}

Figure \ref{fig:face_recoloring} shows examples of image recoloring using our ReHistoGAN. A comparison with the recent high-resolution daytime translation (HiDT) method \cite{anokhin2020high} is shown in Fig.\ \ref{fig:comparison_w_HiDT}.
Additional comparisons with image recoloring and style transfer methods are shown in Fig.\ \ref{fig:compariosns_recoloring_class_based}.
Arguably, our ReHistoGAN produces image recoloring results that are visually more compelling than the results of other methods for image color/style transfer.
As shown in Fig.\ \ref{fig:compariosns_recoloring_class_based}, our ReHistoGAN produces realistic recoloring even when the target image is from a different domain than the input image, compared to other image style transfer methods (e.g., \cite{gatys2016image, sheng2018avatar}).

\begin{figure}
\includegraphics[width=\linewidth]{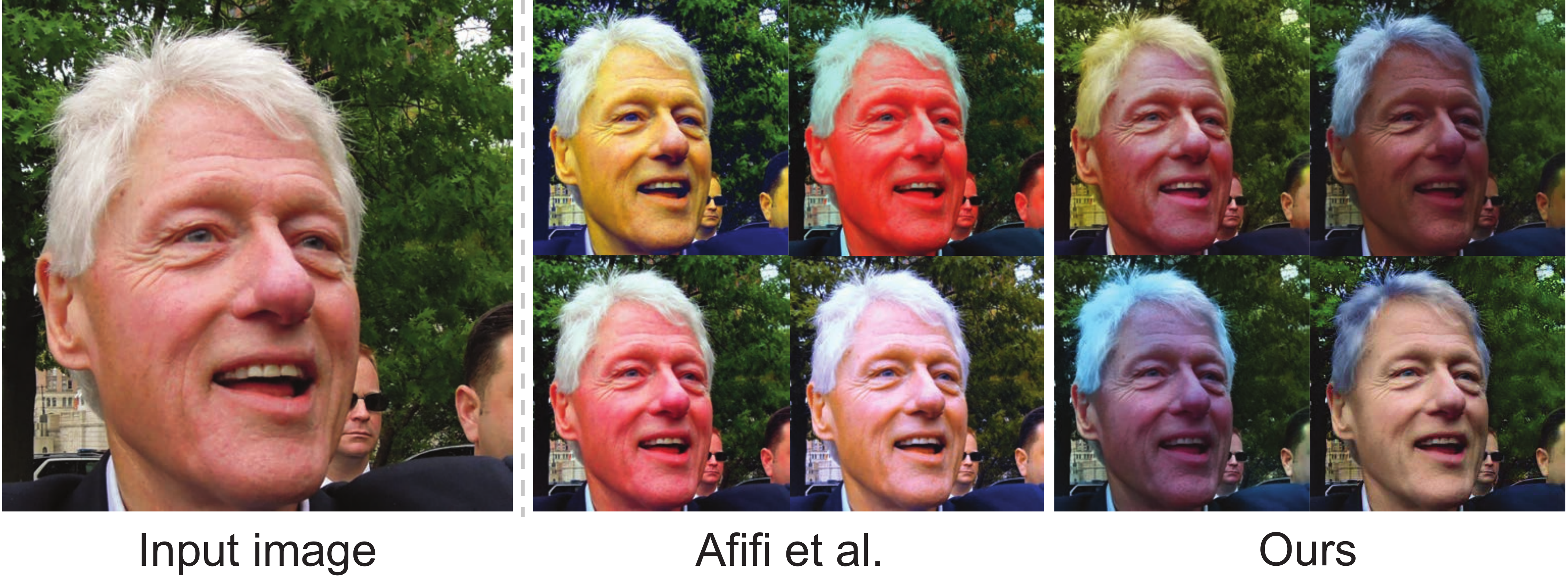}
\vspace{-6.5mm}
\caption{Automatic recoloring comparison with the recent method by Afifi et al., \cite{afifi2019image}.}
\label{fig:compariosns_auto_recoloring}
\end{figure}

Lastly, we provide a qualitative comparison with the recent auto-recoloring method proposed by Afifi et al., \cite{afifi2019image} in Fig.\ \ref{fig:compariosns_auto_recoloring}.
In the shown example, our target histograms were dynamically generated by sampling from a pre-defined set of histograms and applying a linear interpolation between the sampled histograms (see Eq.\ \ref{eq.target_hist}).

\begin{figure}
\centering
\includegraphics[width=\linewidth]{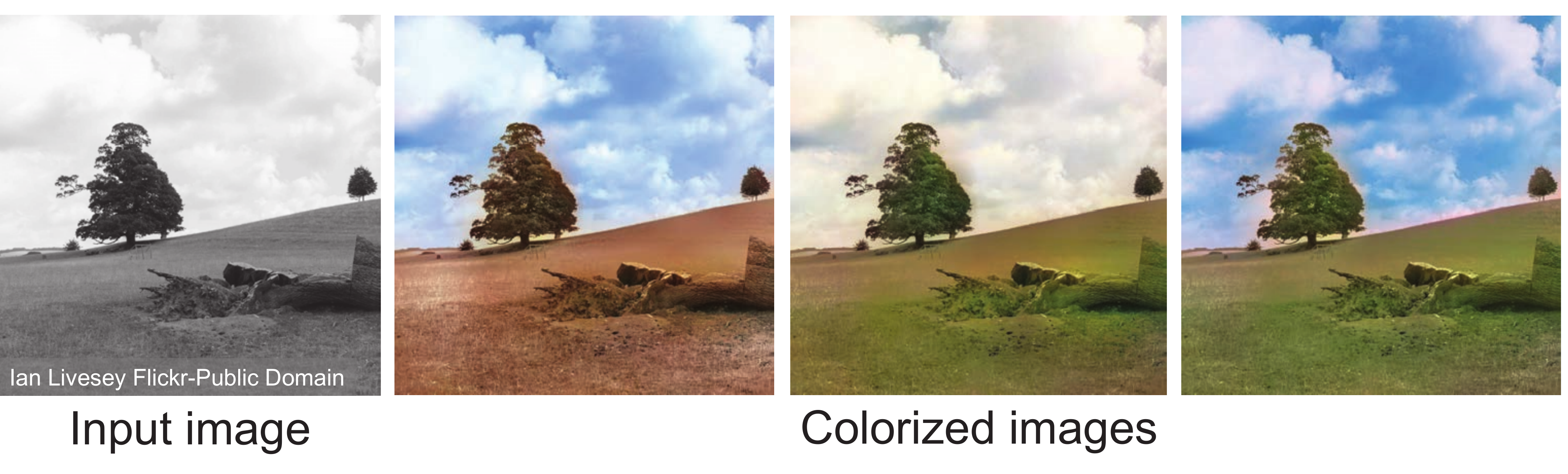}
\vspace{-6.5mm}
\caption{Results of using our ReHistoGAN for a diverse image colorization.}
\label{fig:colorization}
\end{figure}

\vspace{-1mm}
\paragraph{What is Learned?}
Our method learns to map color information, represented by the target color histogram, to an output image's colors with a realism consideration in the recolored image.  Maintaining realistic results is achieved by learning proper matching between the target colors and the input image's semantic objects (e.g., grass can be green, but not blue). To demonstrate this, we examine a trained ReHistoGAN model for an image colorization task, where the input image is grayscale.  The input of a grayscale image means that our ReHistoGAN model has no information regarding objects' colors in the input image.  Figure\ \ref{fig:colorization} shows outputs where the input has been ``colorized''.  As can be seen, the output images have been colorized with good semantic-color matching based on the image's content.

\section{Conclusion} \label{sec.conclusion}

We have presented HistoGAN, a simple, yet effective, method for controlling colors of GAN-generated images.
Our HistoGAN framework learns how to transfer the color information encapsulated in a target histogram feature to the colors of a generated output image.\
To the best of our knowledge, this is the first work to control the color of GAN-generated images directly from color histograms.
Color histograms provide an abstract representation of image color that is decoupled from spatial information.
This allows the histogram representation to be less restrictive and suitable for GAN-generation across arbitrary domains.

We have shown that HistoGAN can be extended to control colors of real images in the form of the ReHistoGAN model.
Our recoloring results are visually more compelling than currently available solutions for image recoloring.
Our image recoloring also enables ``auto-recoloring'' by sampling from a pre-defined set of histograms.
This allows an image to be recolored to a wide range of visually plausible variations.
HistoGAN can serve as a step towards intuitive color control for GAN-based graphic design and artistic endeavors.

\newcommand{\beginsupplement}{%
        \setcounter{table}{0}
        \renewcommand{\thetable}{S\arabic{table}}%
        \setcounter{figure}{0}
        \renewcommand{\thefigure}{S\arabic{figure}}%
     }

\section{Supplementary Material}
\beginsupplement

\subsection{Details of Our Networks} \label{sec:network}

\begin{figure}[b]
\centering
\includegraphics[width=0.72\linewidth]{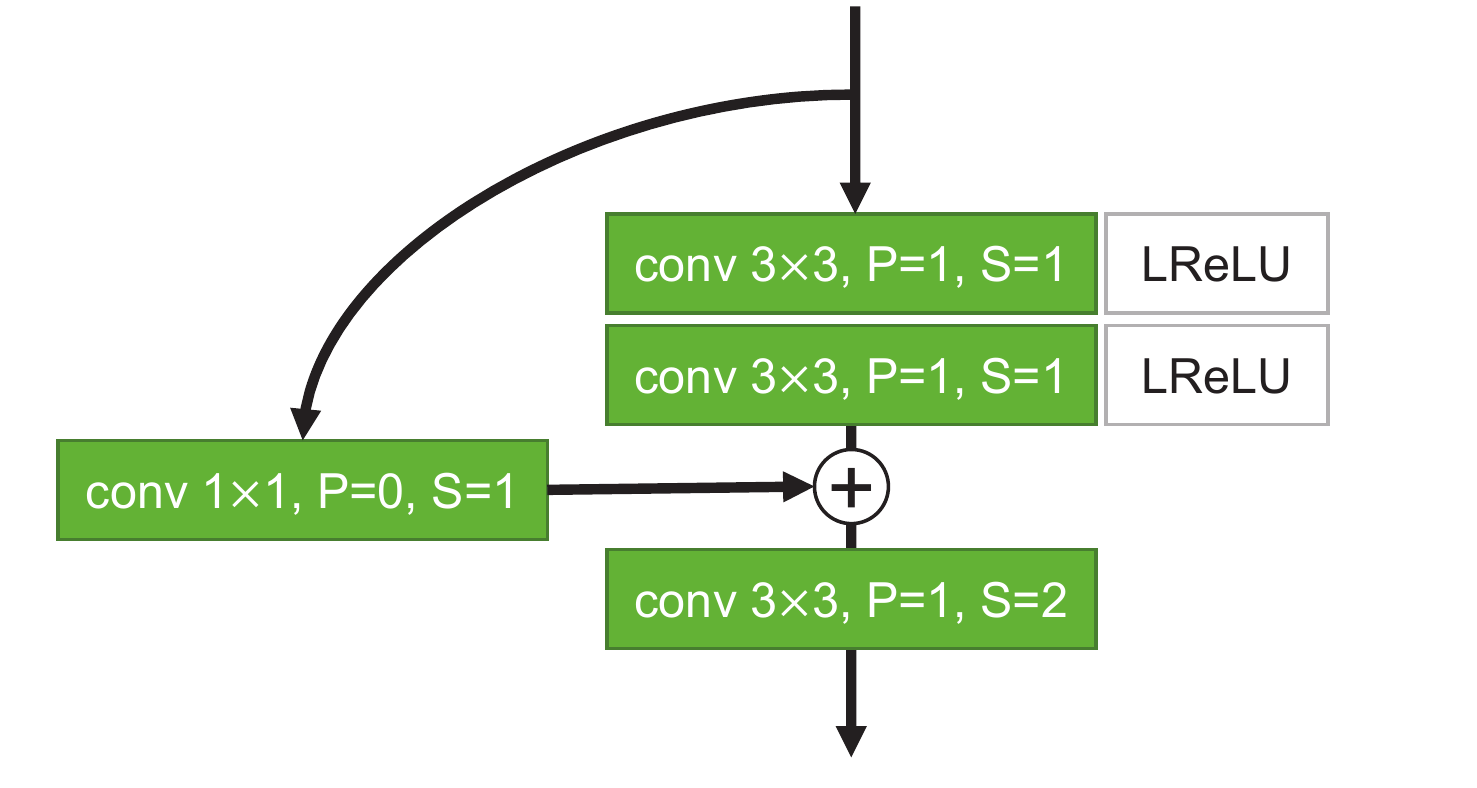}
\vspace{-2mm}
\caption{Details of the residual discriminator block used to reconstruct our discriminator network. The term P and S refer to the padding and stride used in each layer.}\vspace{-2mm}
\label{fig:discriminator_block}
\end{figure}

Our discriminator network, used in all of our experiments, consists of a sequence of $\log_2(N) – 1$ residual blocks, where $N$ is the image width/height, and the last layer is an fully connected (fc) layer that produces a scalar feature. The first block accepts a three-channel input image and produce $m$ output channels. Then, each block $i$ produces $2m_{i-1}$ output channels (i.e., duplicate the number of output channels of the previous block). The details of the residual blocks used to build our discriminator network are shown in Fig.\ \ref{fig:discriminator_block}.

\begin{figure}
\centering
\includegraphics[width=\linewidth]{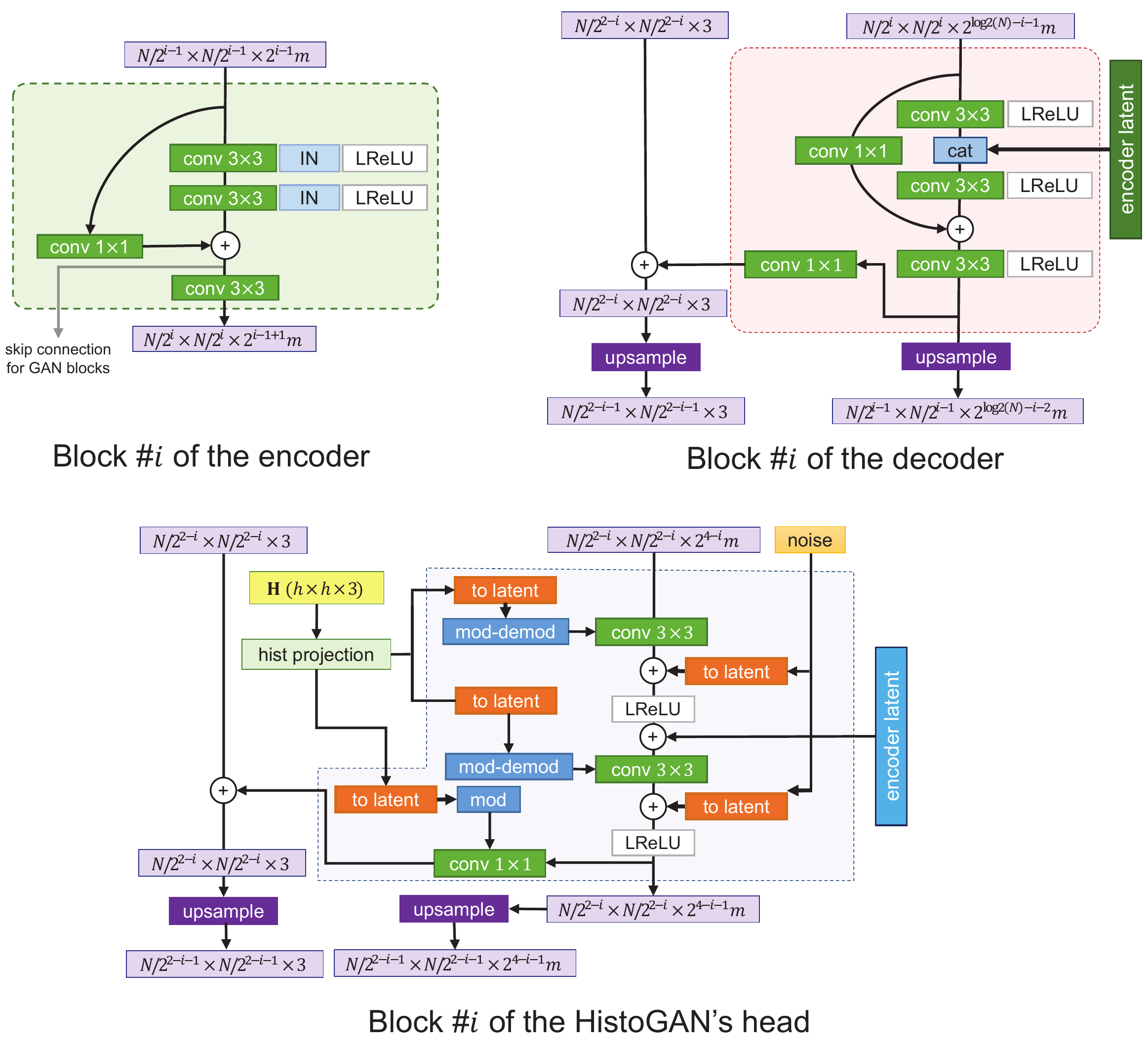}
\vspace{-5.5mm}
\caption{Details of our ReHistoGAN network. We modified the last two blocks of our HistoGAN by adding a gate for the processed skipped features from the first two blocks of our encoder.}\vspace{-2mm}
\label{fig:recoloring-design_}
\end{figure}

Figure~\ref{fig:recoloring-design_} provides the details of our encoder, decoder and GAN blocks used in our ReHistoGAN (used for image recoloring). As shown, we modified the last two blocks of our HistoGAN's to accept the latent feature passed from the first two blocks of our encoder. This modification helps our HistoGAN's head to consider both information of the input image structure and the target histogram in the recoloring process.

\subsection{Training Details} \label{sec:training}

We train our networks using an NVIDIA TITAN X (Pascal) GPU. For HistoGAN training, we optimized both the generator and discriminator networks using the diffGrad optimizer \cite{8939562}. In all experiments, we set the histogram bin, $h$, to 64 and the fall-off parameter of our histogram's bins, $\tau$, was set to 0.02. We adopted the exponential moving average of generator network's weights \cite{karras2019style, karras2020analyzing} with the path length penalty, introduced in StyleGAN \cite{karras2020analyzing}, every 32 iterations to train our generator network. Due to the hardware limitation, we used mini-batch of 2 with accumulated gradients every 16 iteration steps and we set the image's dimension, $N$, to 256. We set the scale factor of the Hellinger distance loss, $\alpha$, to 2 (see Sec.\ \ref{sec:ablations} for an ablation study).

\begin{figure}
\centering
\includegraphics[width=\linewidth]{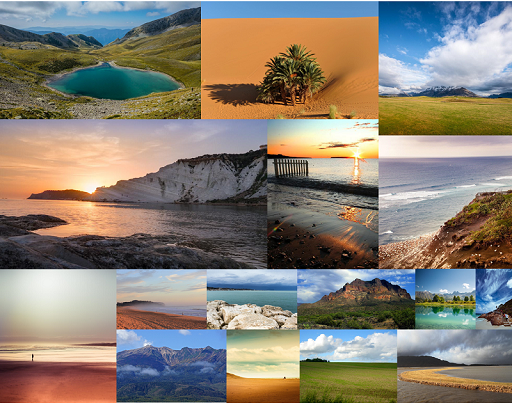}
\vspace{-5.5mm}
\caption{Examples taken from our set of 4,316 landscape images collected from Flickr.}\vspace{-2mm}
\label{fig:landscape_dataset}
\end{figure}

As mentioned in the main paper, we trained our HistoGAN using several domain datasets, including: human faces \cite{karras2019style}, flowers \cite{nilsback2008automated}, cats \cite{catdataset}, dogs \cite{khosla2011novel}, birds \cite{wah2011caltech}, anime faces \cite{animedataset}, human hands \cite{afifi201911k}, bedrooms \cite{yu2015lsun}, cars \cite{krause20133d}, and aerial scenes \cite{maggiori2017can}. We further trained our HistoGAN using 4,316 landscape images collected from Flickr. The collected images have one of the following copyright licenses: no known copyright restrictions, Public Domain Dedication (CC0), or Public Domain Mark. See Fig.\ \ref{fig:landscape_dataset} for representative examples from the landscape set.

To train our ReHistoGAN, we used the diffGrad optimizer \cite{8939562} with the same mini-batch size used to train our HistoGAN. We trained our network using the following hyperparameters $\alpha=2$, $\beta=1.5$, $\gamma=32$ for 100,000 iterations. Then, we continued training using $\alpha=2$, $\beta=1$, $\gamma=8$ for additional 30,000 iterations to reduce potential artifacts in recoloring (see Sec.\ \ref{sec:ablations} for an ablation study).

\subsection{Ablation Studies} \label{sec:ablations}

We carried out a set of ablation experiments to study the effect of different values of hyperparameters used in the main paper. Additionally, we show results obtained by variations in our loss terms.

\begin{figure}
\centering
\includegraphics[width=\linewidth]{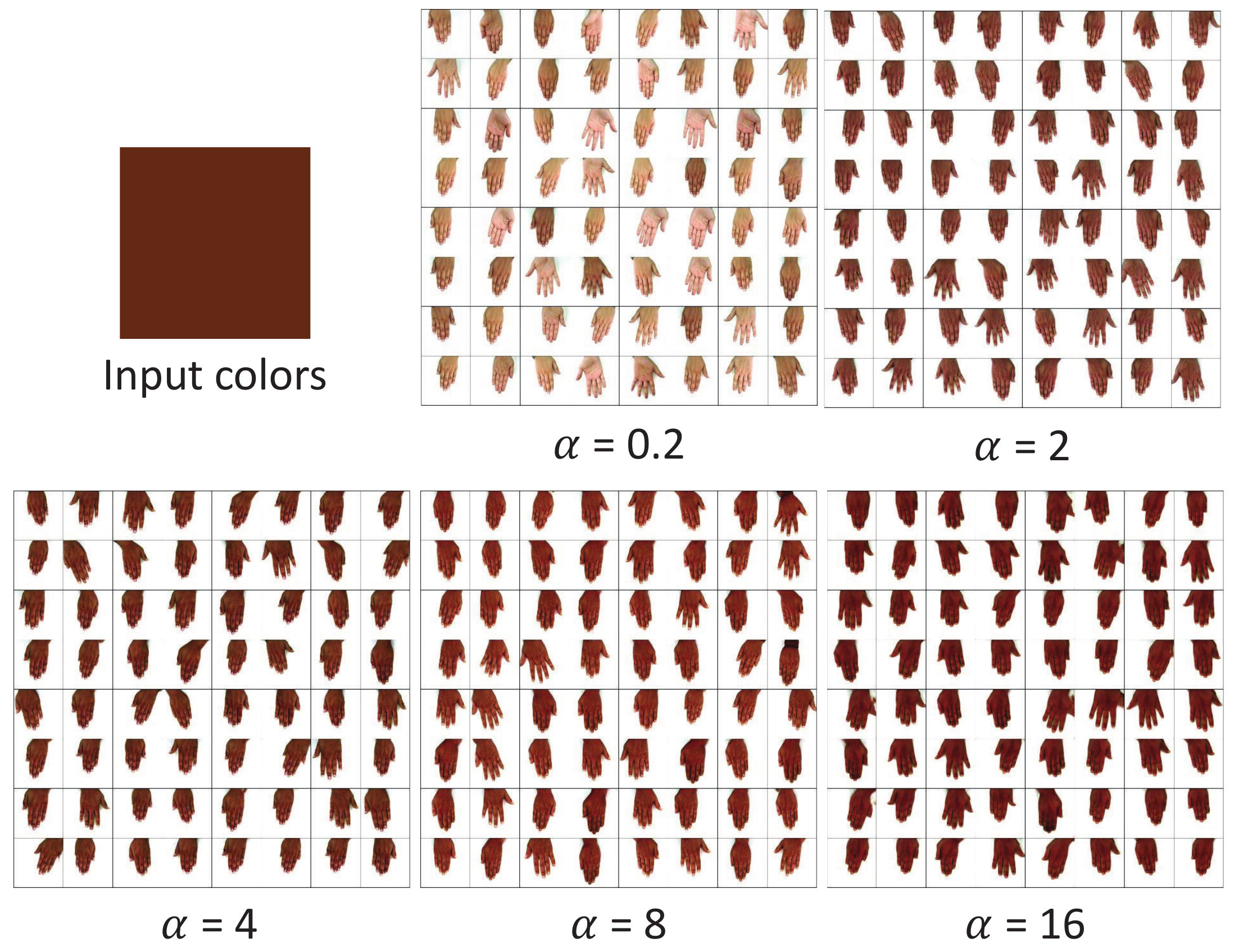}
\vspace{-5.5mm}
\caption{Results obtained by training our HistoGAN in hand images \cite{afifi201911k} using different values of $\alpha$.}\vspace{-2mm}
\label{fig:alpha_ablation}
\end{figure}

We begin by studying the effect of the scale factor, $\alpha$, used in the loss function to train our HistoGAN. This scale factor was used to control strength of the histogram loss term. In this set of experiments, we used the 11K Hands dataset \cite{afifi201911k} to be our target domain and trained our HistoGAN with the following values of $\alpha$: 0.2, 2, 4, 8, and 16. Table \ref{table:ablation_results} shows the evaluation results using the Frech\'et inception distance (FID) metric \cite{heusel2017gans}, the KL divergence, and Hellinger distance. The KL divergence and Hellinger distance were used to measure the similarity between the target histogram and the histogram of GAN-generated images. Qualitative comparisons are shown in Fig.\ \ref{fig:alpha_ablation}

\begin{table}[]
\centering
\caption{Results of our HistoGAN using different values of $\alpha$. In this set of experiments, we used the Hands dataset \cite{afifi201911k} as our target domain. The term FID stands for the Frech\'et inception distance metric \cite{heusel2017gans}. The term KL Div. refers to the KL divergence between the histograms of the input image and generated image, while the term H. dis. refers to Hellinger distance.\label{table:ablation_results}}
\scalebox{0.87}{
\begin{tabular}{|c|c|c|c|}
\hline
 &  & \multicolumn{2}{c|}{RGB-$uv$ hist.} \\ \cline{3-4}
\multirow{-2}{*}{$\alpha$} & \multirow{-2}{*}{FID} & KL Div. & H dist. \\ \hline
0.2 & 1.9950 & 0.3935 & 0.3207 \\ \hline
\rowcolor[HTML]{FFFFC7}
2 & \textbf{2.2438} & \textbf{0.0533} & \textbf{0.1085} \\ \hline
4 & 6.8750 & 0.0408 & 0.0956 \\ \hline
8 & 9.4101 & 0.0296 & 0.0822 \\ \hline
16 & 15.747 & 0.0237 & 0.0743 \\ \hline
\end{tabular}}
\end{table}

\begin{figure*}
\centering
\includegraphics[width=\linewidth]{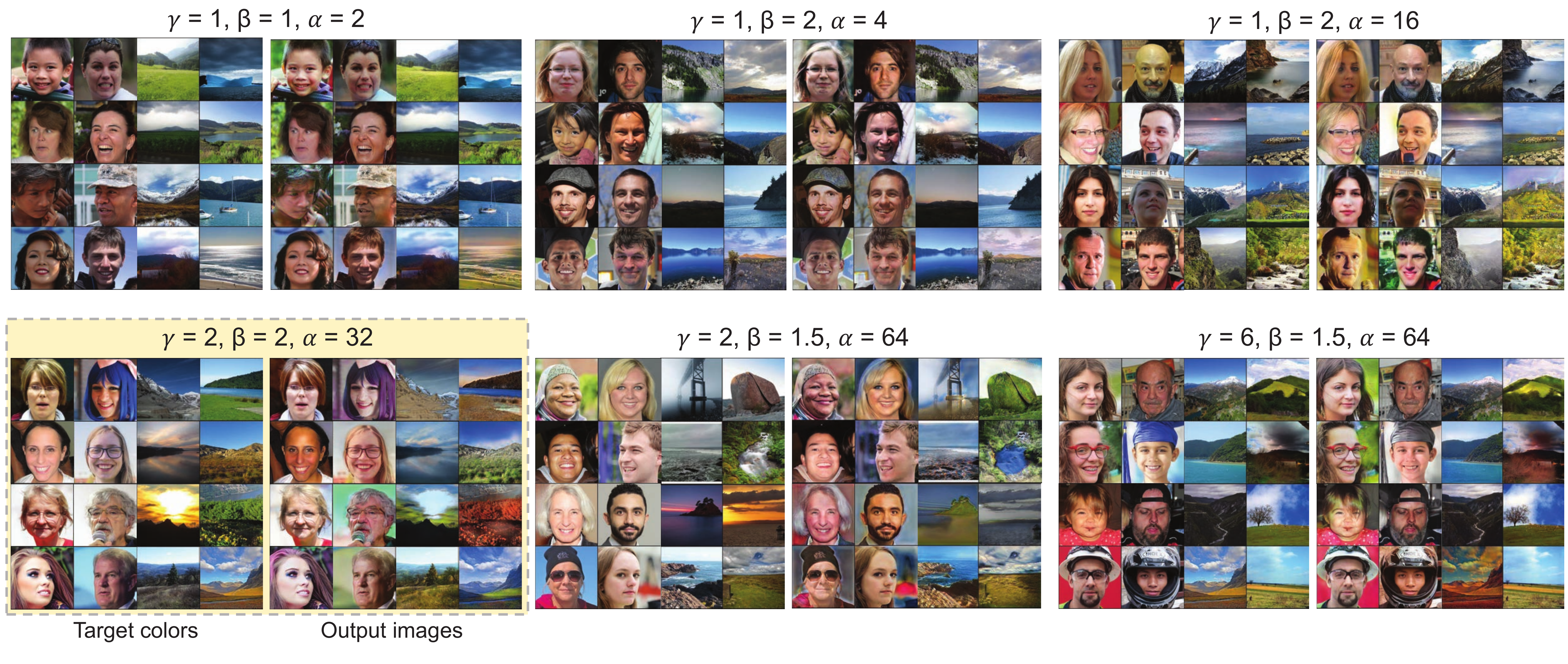}
\vspace{-6.5mm}
\caption{Results of recoloring by training our recoloring network using different values of $\alpha$, $\beta$, and $\gamma$ hyperparameters. The highlighted results refer to the settings used to produce the reported results in the main paper and the supplementary materials.}\vspace{-2mm}
\label{fig:alpha_beta_gamma_ablation}
\end{figure*}

Figure \ref{fig:alpha_beta_gamma_ablation} shows examples of recoloring results obtained by trained ReHistoGAN models using different combination values of $\alpha$, $\beta$, $\gamma$. As can be seen, a lower value of the scale factor, $\alpha$, of the histogram loss term results in ignoring our network to the target colors, while higher values of the scale factor, $\gamma$, of the discriminator loss term, make our method too fixated on producing realistic output images, regardless of achieving the recoloring (i.e., tending to re-produce the input image as is).

\begin{figure}
\includegraphics[width=\linewidth]{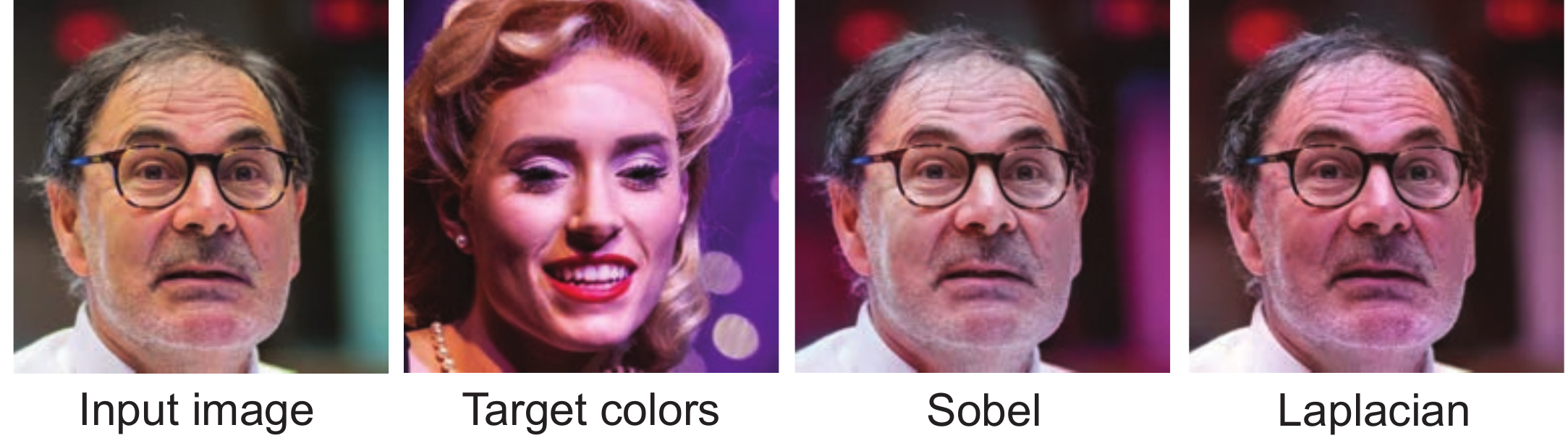}
\vspace{-5.5mm}
\caption{Results of two different kernels used to compute the reconstruction loss term.}\vspace{-2mm}
\label{fig:sobel_vs_laplacian}
\end{figure}

In the recoloring loss, we used a reconstruction loss term to retain the input image's spatial details in the output recolored image. Our reconstruction loss is based on the derivative of the input image. We have examined two different kernels, which are: the vertical and horizontal $3\!\times\!3$ Sobel kernels (i.e., the first-order derivative approximation) and the $3\!\times\!3$ Laplacian kernel (i.e., the second-order derivative). We found that training using both kernels give reasonably good results, while the Laplacian kernel produces more compiling results in most cases; see Fig.\ \ref{fig:sobel_vs_laplacian} for an example.

\begin{figure}[b]
\includegraphics[width=\linewidth]{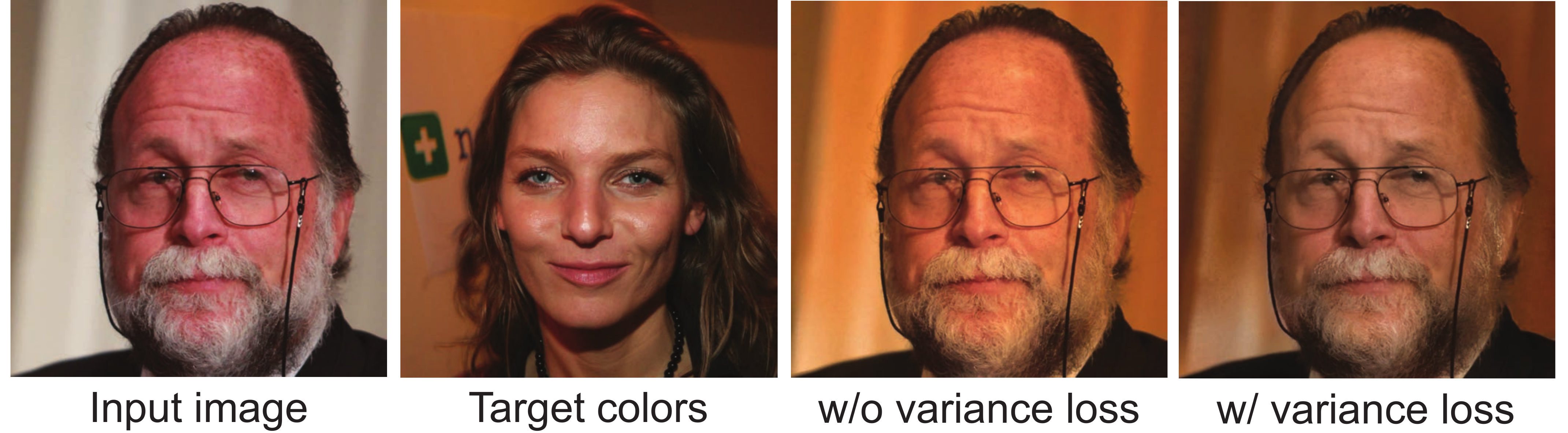}
\vspace{-5.5mm}
\caption{The impact of the variance loss term. The shown results were obtained by training our ReHistoGAN with and without the variance loss term.}\vspace{-2mm}
\label{fig:variance_loss_supp}
\end{figure}

In the main paper, we introduced a variance loss term to encourage our network to avoid the global color cast solution for image recoloring. Figure \ref{fig:variance_loss_supp} shows an example of the global color cast problem, where the network applies a global color shift to the input image to match the target histogram. As shown in  Fig.\ \ref{fig:variance_loss_supp} after training our network with the variance loss, this problem is reduced.

\subsection{Universal ReHistoGAN Model}\label{sec:universal}

As the case of most GAN methods, our ReHistoGAN targets a specific object domain to achieve the image recoloring task. This restriction may hinder the generalization of our method to deal with images taken from arbitrary domains. To deal with that, we collected images from a different domain, aiming to represent the ``universal'' object domain.

Specifically, our training set of images contains $\sim$2.4 million images collected from different image datasets. These datasets are: collection from the Open Images dataset \cite{kuznetsova2020open}, the MIT-Adobe FiveK dataset \cite{fivek}, the Microsoft COCO dataset \cite{lin2014microsoft}, the CelebA dataset \cite{liu2015deep}, the Caltech-UCSD birds-200-2011 dataset \cite{wah2011caltech}, the Cats dataset \cite{catdataset}, the Dogs dataset \cite{khosla2011novel}, the Cars dataset \cite{krause20133d}, the OxFord Flowers dataset \cite{nilsback2008automated}, the LSUN dataset \cite{yu2015lsun}, the ADE20K dataset \cite{zhou2017scene, zhou2019semantic}, and the FFHQ dataset \cite{karras2019style}. We also added Flickr images collected using the following keywords: $\texttt{landscape}$, $\texttt{people}$, $\texttt{person}$, $\texttt{portrait}$, $\texttt{field}$, $\texttt{city}$, $\texttt{sunset}$, $\texttt{beach}$, $\texttt{animals}$, $\texttt{living room}$, $\texttt{home}$, $\texttt{house}$, $\texttt{night}$, $\texttt{street}$, $\texttt{desert}$, $\texttt{food}$. We have excluded any grayscale image from the collected image set.

\begin{figure}
\centering
\includegraphics[width=\linewidth]{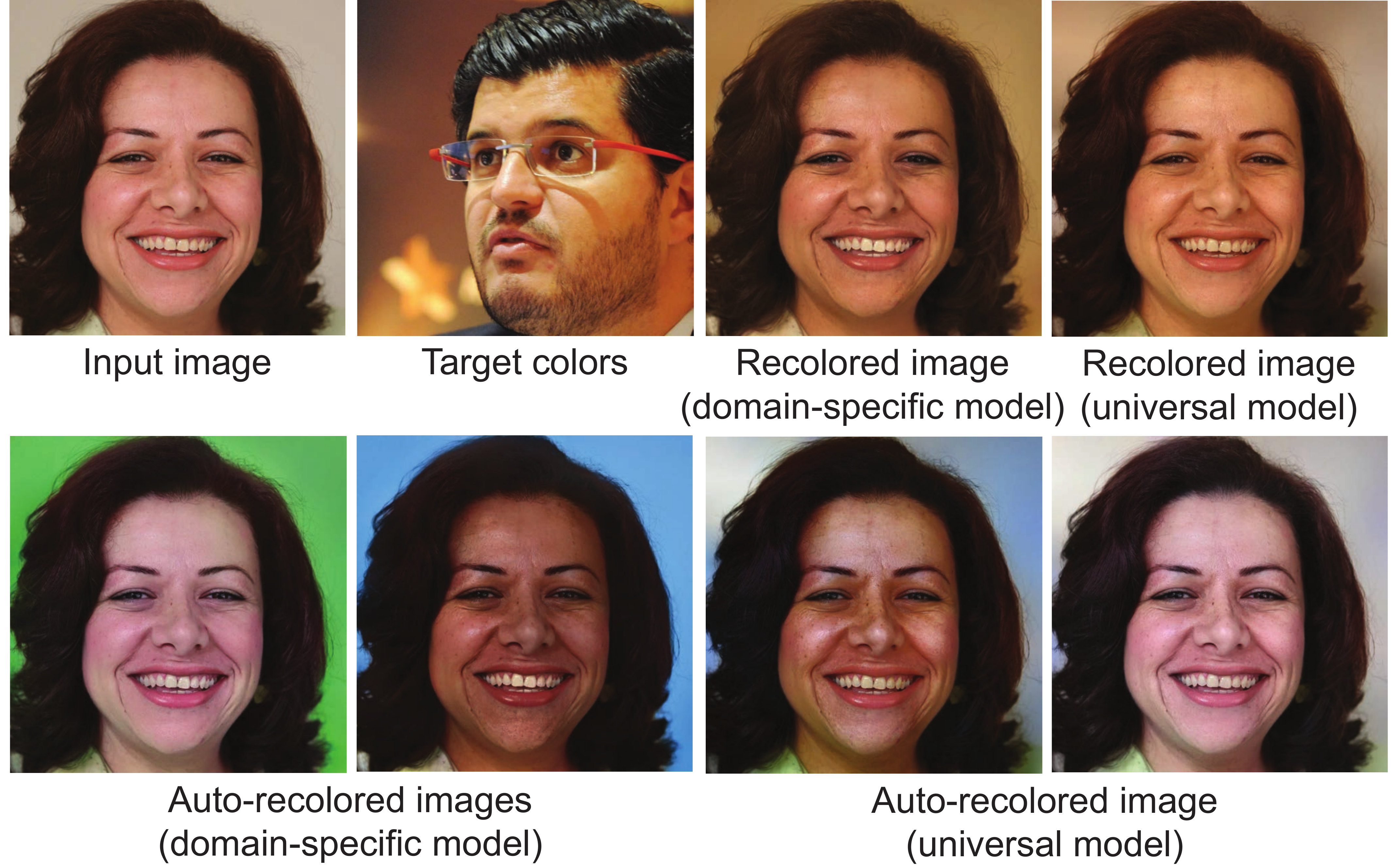}
\vspace{-5.5mm}
\caption{Results of domain-specific and universal ReHistoGAN models. We show results of using a given target histogram for recoloring and two examples of the auto recoloring results of each model.}\vspace{-2mm}
\label{fig:object_specific_vs_universal_recoloring}
\end{figure}

\begin{figure}[b]
\centering
\includegraphics[width=\linewidth]{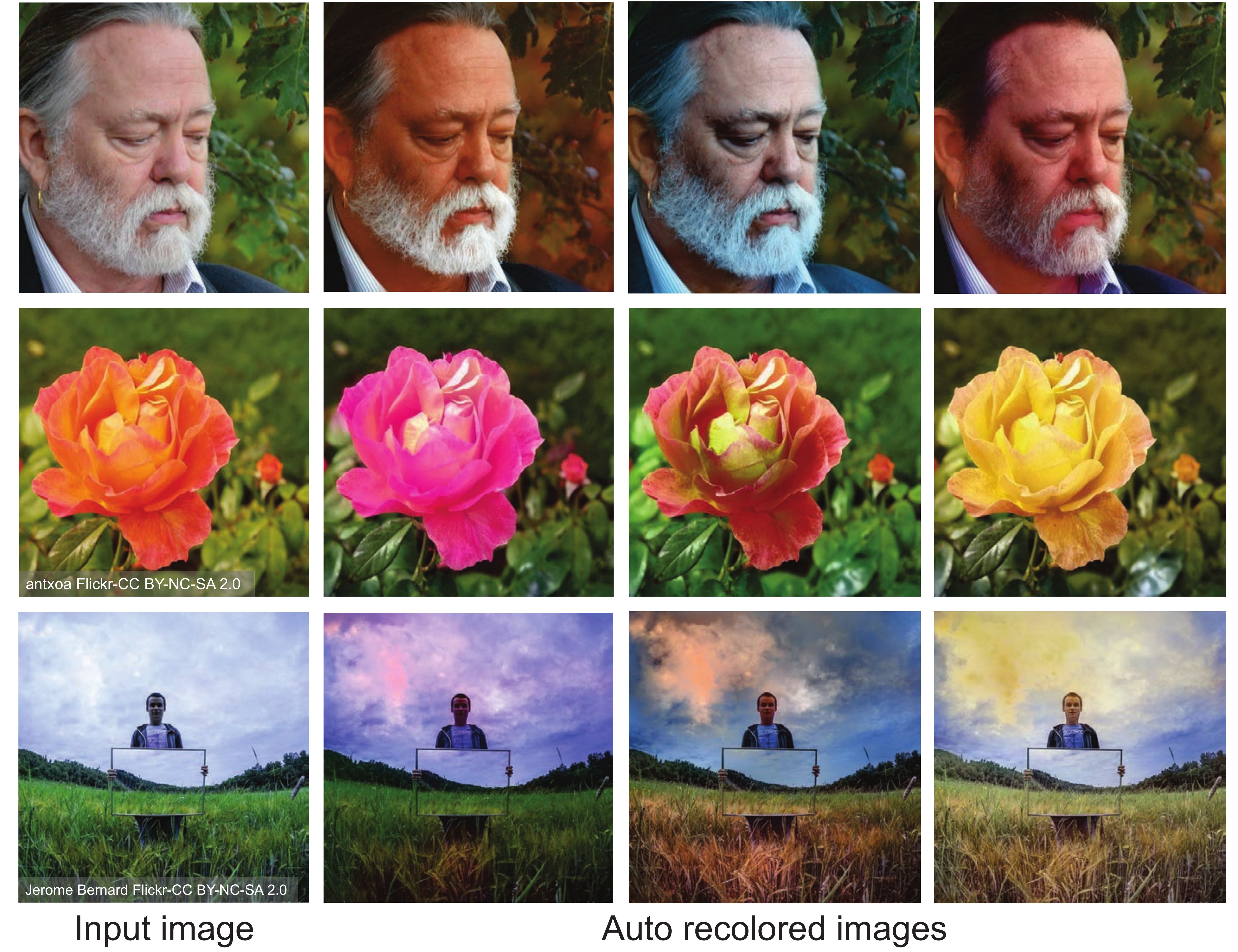}
\vspace{-5.5mm}
\caption{Auto recoloring using our universal ReHistoGAN model.}\vspace{-2mm}
\label{fig:universal_model_auto_results}
\end{figure}

We trained our ``universal'' model using $m=18$ on this collected set of 2,402,006 images from several domains. The diffGrad optimizer \cite{8939562} was used to minimize the same generator loss described in the main paper using the following hyperparameters $\alpha=2$, $\beta=1.5$, $\gamma=32$ for 150,000 iterations. Then, we used $\alpha=2$, $\beta=1$, $\gamma=8$ to train the model for additional 350,000 iterations. We set the mini-batch size to 8 with an accumulated gradient every 24 iterations. Figure \ref{fig:object_specific_vs_universal_recoloring} show results of our domain-specific and universal models for image recoloring. As can be seen, both models produce realistic recoloring, though the universal model tends to produce recolored images with less vivid colors compared to our domain-specific model. Additional examples of auto recoloring using our universal model are shown in Fig. \ref{fig:universal_model_auto_results}.

\subsection{Limitations} \label{sec:limitations}

\begin{figure}
\centering
\includegraphics[width=\linewidth]{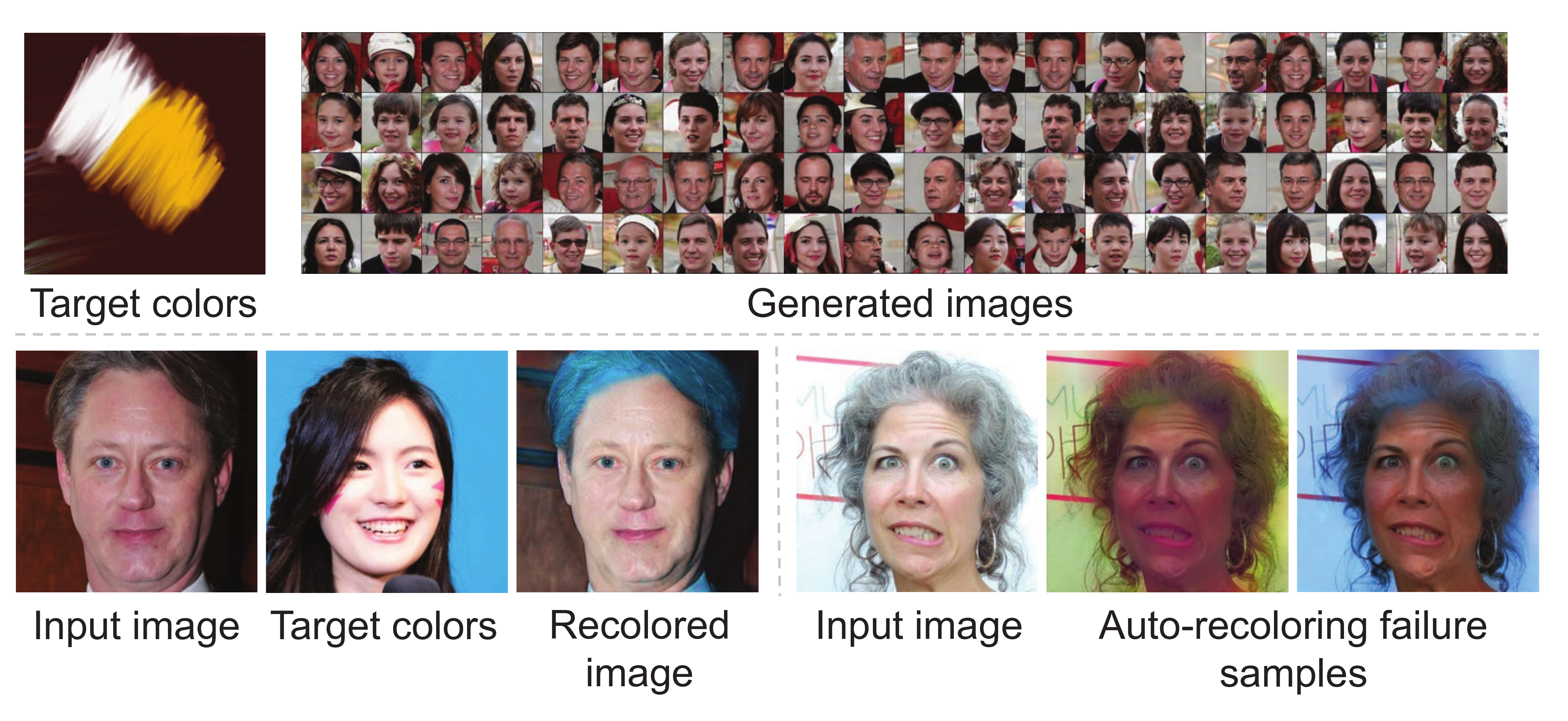}
\vspace{-5.5mm}
\caption{Failure cases of HistoGAN and ReHistoGAN. Our HistoGAN fails sometimes to consider all colors of target histogram in the generated image. Color bleeding is another problem that could occur in ReHistoGAN's results, where our network could not properly allocate the target (or sampled) histogram colors in the recolored image.}\vspace{-2mm}
\label{fig:failure_cases}
\end{figure}

\begin{figure}
\includegraphics[width=\linewidth]{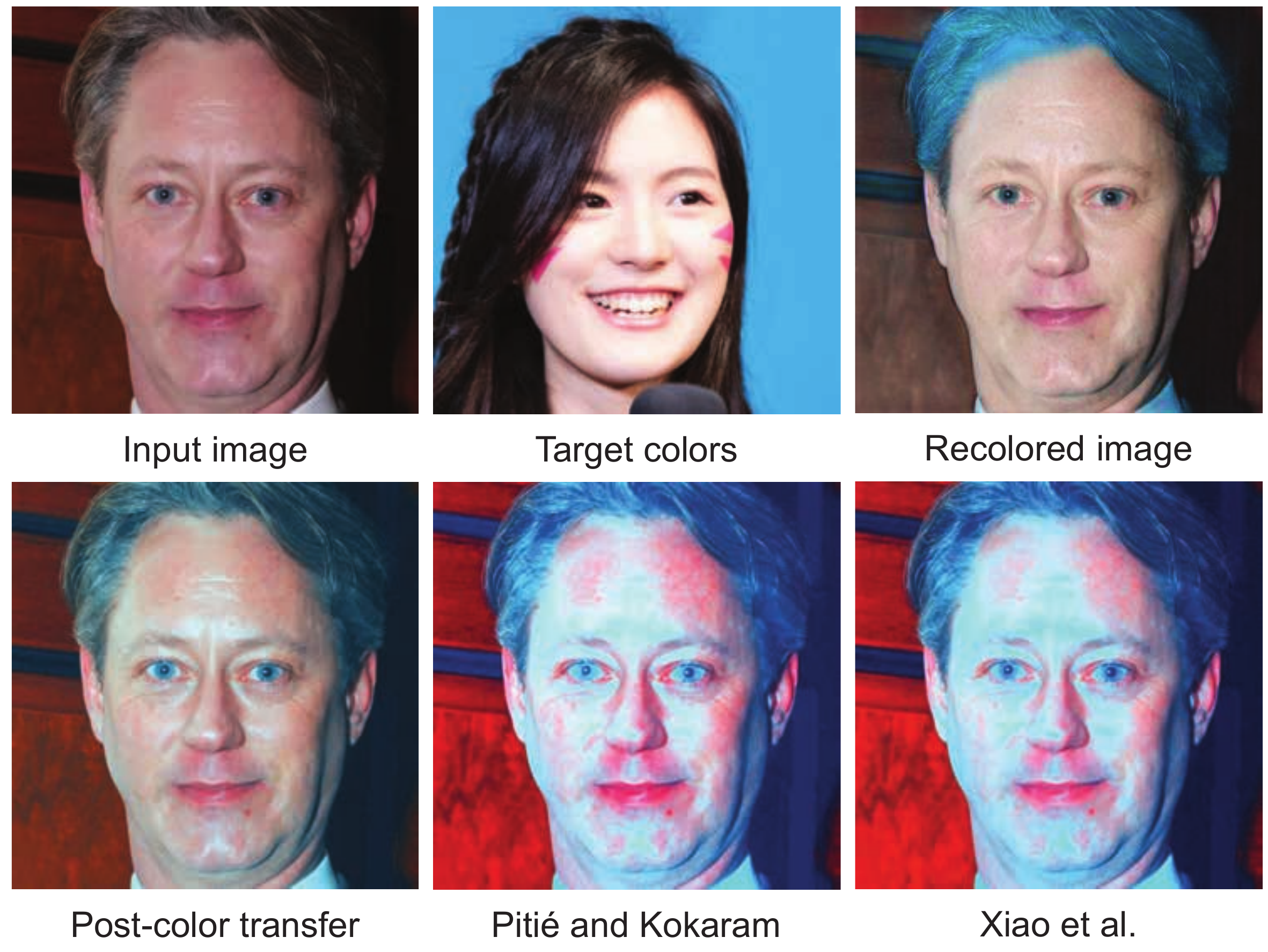}
\vspace{-4mm}
\caption{To reduce potential color bleeding artifacts, it is possible to apply a post-color transfer to our initial recolored image colors to the input image. The results of adopting this strategy are better than applying the color transfer to the input image in the first place. Here, we use the color transfer method proposed by Piti\'e and Kokaram \cite{pitie2007} as our post-color transfer method. We also show the results of directly applying Piti\'e and Kokaram's \cite{pitie2007} method to the input image.}\vspace{-2mm}
\label{fig:fixing_failure_cases}
\end{figure}

\begin{figure}
\includegraphics[width=\linewidth]{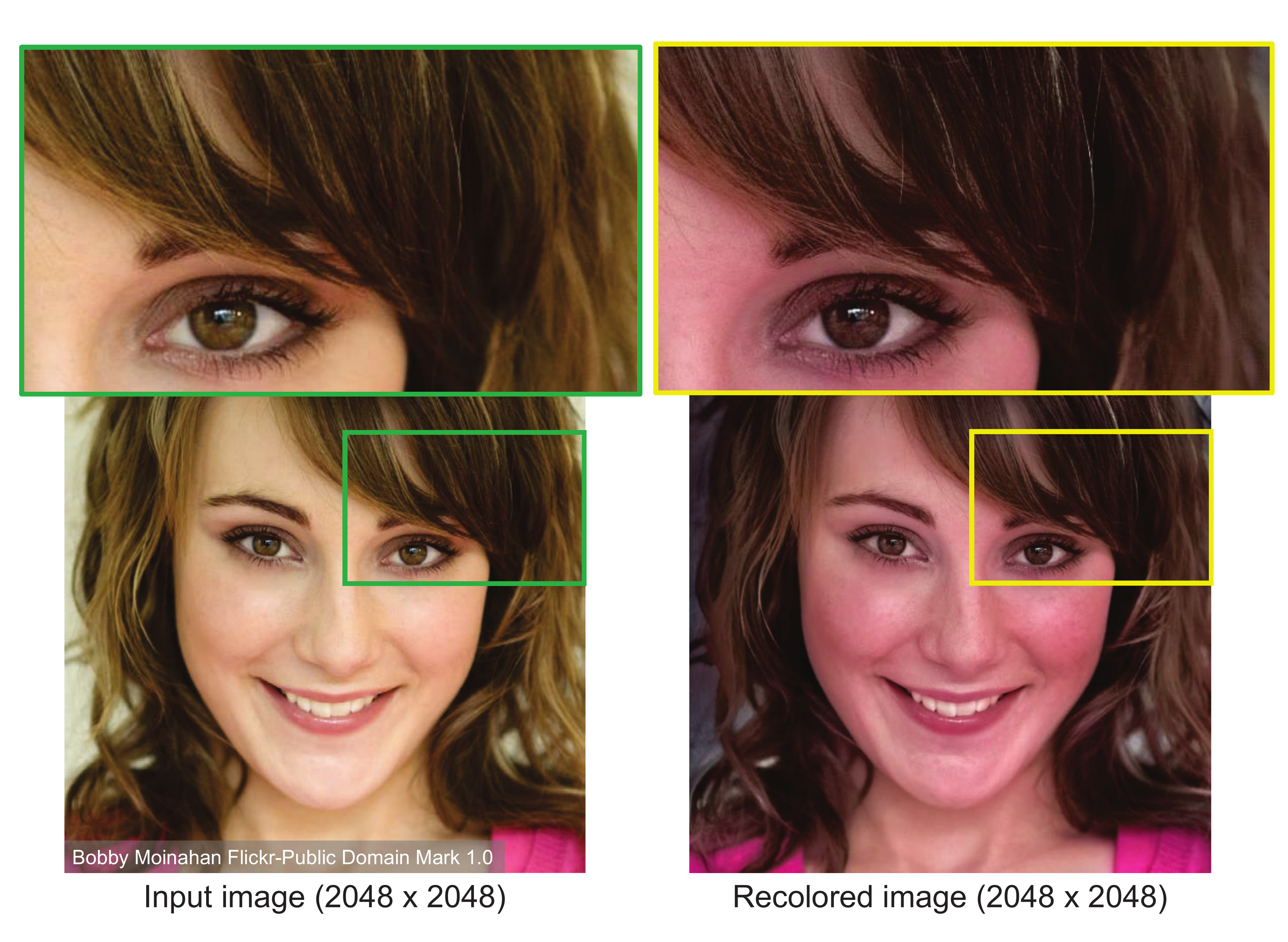}
\vspace{-5.5mm}
\caption{We apply the bilateral guided upsampling \cite{chen2016bilateral} as a post-processing to reduce potential artifacts of dealing with high-resolution images in the inference phase. In the shown example, we show our results of recoloring using an input image with $2048\!\times\!2048$ pixels.}\vspace{-2mm}
\label{fig:dealing_with_any_size}
\end{figure}

Our method fails in some cases, where the trained HistoGAN could not properly extract the target color information represented in the histogram feature. This problem is due to the inherent limitation of the 2D projected representation of the original target color distribution, where different colors are mapped to the same chromaticity value in the projected space. This is shown in Fig.\ \ref{fig:failure_cases}-top, where the GAN-generated images do not have all colors in the given target histogram. Another failure case can occur in image recoloring, where the recolored images could have some color-bleeding artifacts due to errors in allocating the target/sampled histogram colors in the recolored image. This can be shown in Fig. \ref{fig:failure_cases}-bottom

\begin{figure}
\includegraphics[width=\linewidth]{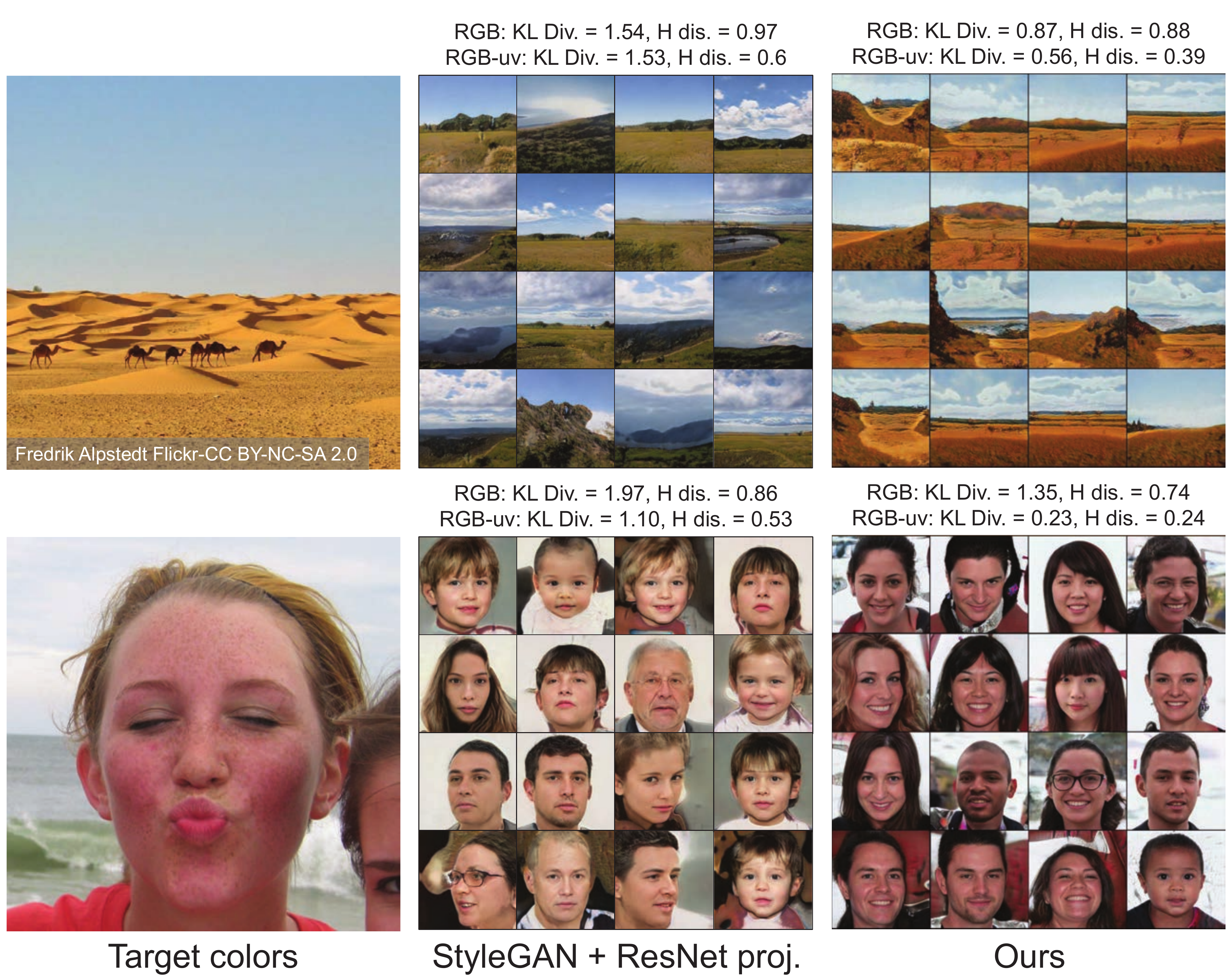}
\vspace{-5.5mm}
\caption{Comparison with generated images using StyleGAN \cite{karras2020analyzing} with latent space projection (see the main paper for more details) and our results.}\vspace{-2mm}
\label{fig:compariosn_w_styleGAN}
\end{figure}

\begin{figure}
\includegraphics[width=\linewidth]{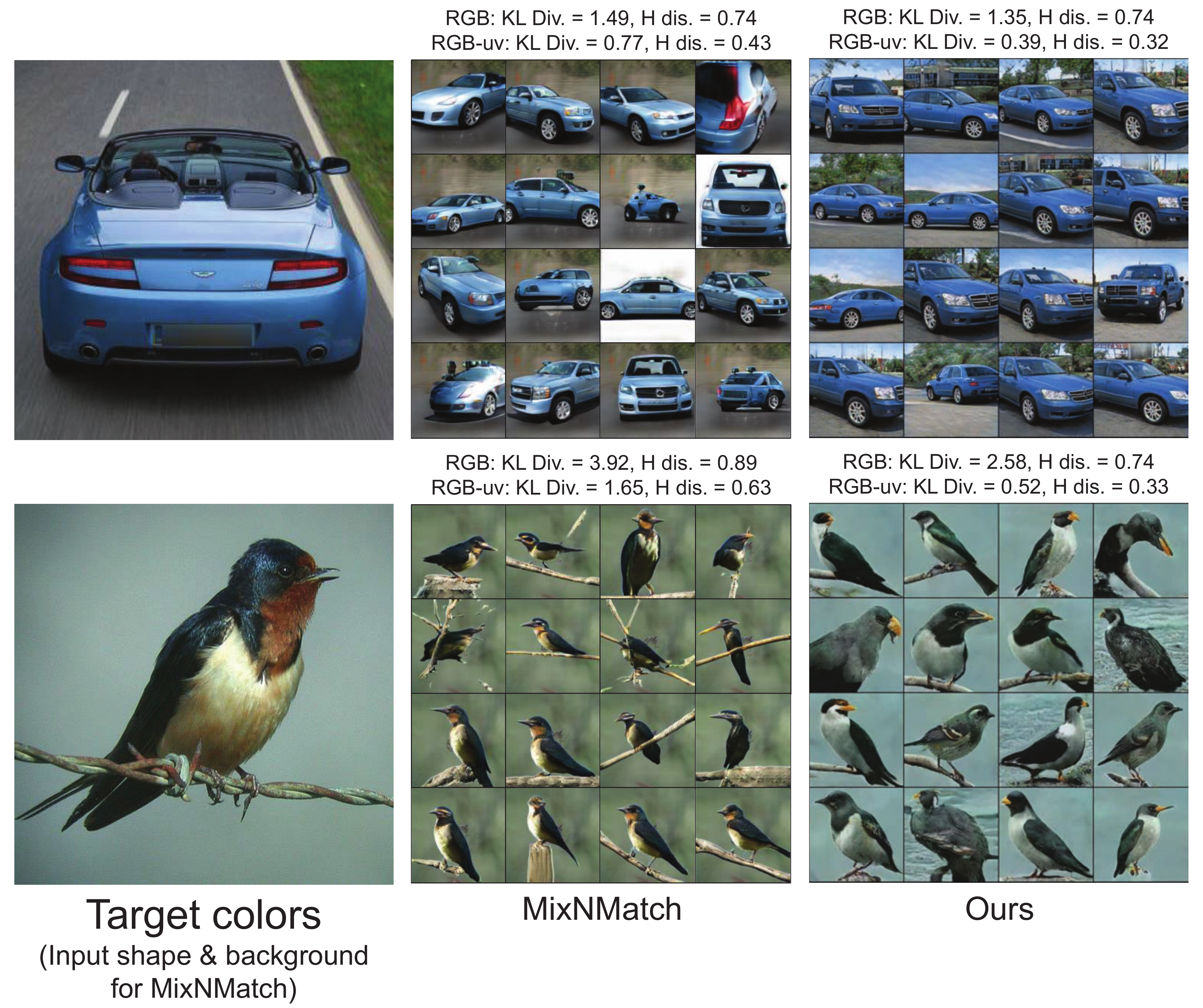}
\vspace{-5.5mm}
\caption{Additional comparison with the MixNMatch method \cite{li2020mixnmatch}. In these examples, the target images were used as input shape and background images for the MixNMatch method.}
\label{fig:comparison_w_MixNMatch_supp}
\end{figure}

\subsection{Post-Processing} \label{sec:post-processing}

As discussed in Sec. \ref{sec:limitations}, our method produces, in some times, results with color bleeding, especially when the target histogram feature has unsuitable color distribution for the content of the input image. This color-bleeding problem can be mitigated using a post-process color transfer between the input image and our initial recoloring. Surprisingly, this post-processing mapping produces results better than adopting the mapping in the first place---namely, applying the color transfer mapping without having our intermediate recoloring result.

\begin{figure*}
\includegraphics[width=\linewidth]{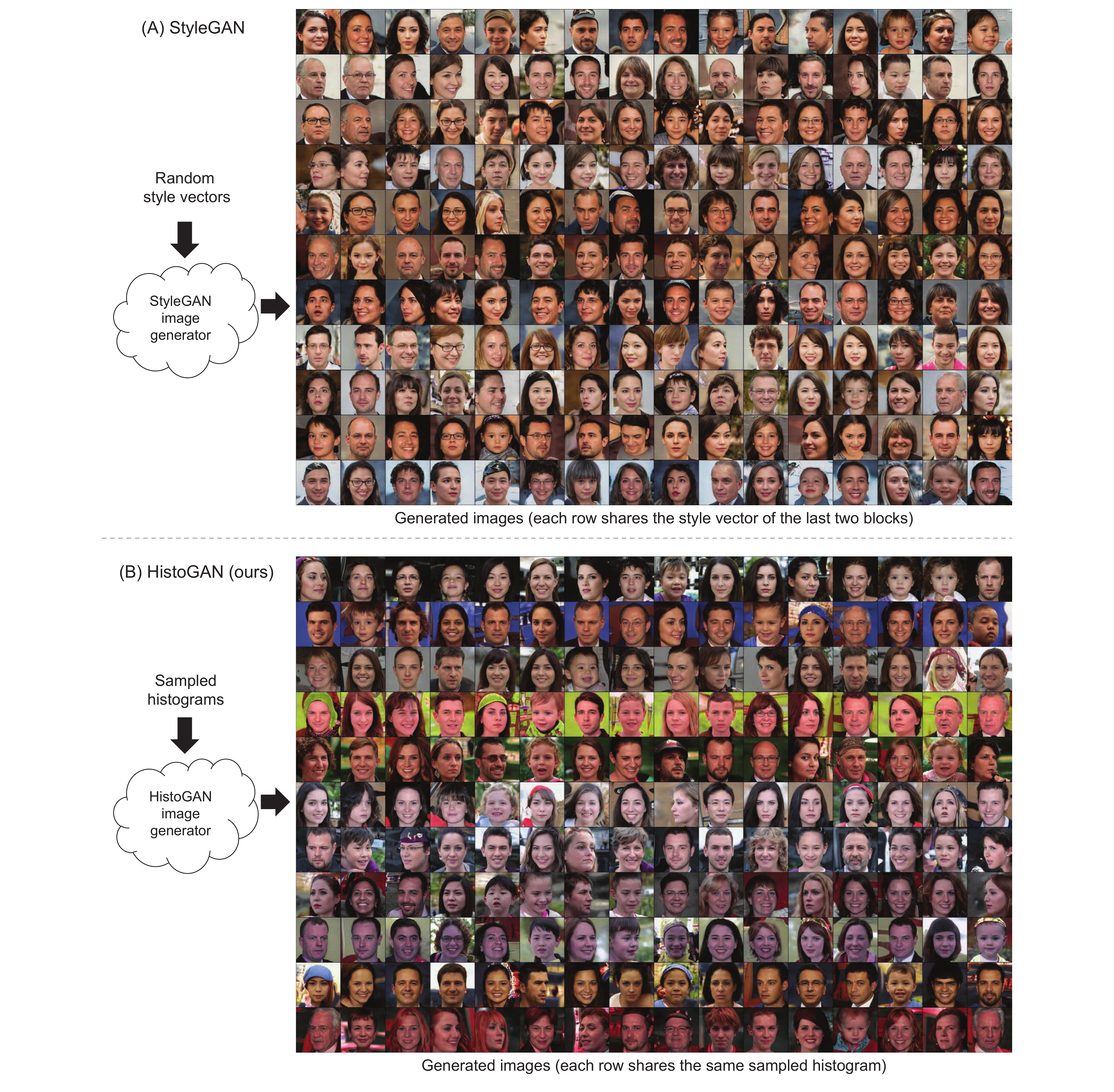}
\vspace{-6.5mm}
\caption{Our HistoGAN can be used to generate ``unlimited'' number of random samples, exactly like traditional StyleGANs \cite{karras2019style, karras2020analyzing}, by sampling from a pre-defined set of histograms to generate target histograms. In the shown figure, we show generated images by StyleGAN \cite{karras2019style, karras2020analyzing} and our HistoGAN. In each row of the StyleGAN-generated images, we fixed the fine-style vector of the last two blocks of the StyleGAN, as these blocks are shown to control the fine-style of the generated image \cite{karras2020analyzing}. We also fixed the generated histogram for each row of our HistoGAN-generated images.}\vspace{-2mm}
\label{fig:histGAN_as_styleGAN_supp}
\end{figure*}

Figure \ref{fig:fixing_failure_cases} shows an example of applying Piti\'e, and Kokaram's method \cite{pitie2007} as a post-processing color transfer to map the colors of the input image to the colors of our recolored image. In the shown figure, we also show the result of using the same color transfer method -- namely, Piti\'e and Kokaram's method \cite{pitie2007} -- to transfer the colors of the input image directly to the colors of the target image. As shown, the result of using our post-process strategy has a better perceptual quality.

Note that except for this figure (i.e., Fig.\ \ref{fig:fixing_failure_cases}), we \textit{did not} adopted this post-processing strategy to produce the reported results in the main paper or the supplementary materials. We discussed it here as a solution to reduce the potential color bleeding problem for completeness.

As our image-recoloring architecture is a fully convolutional network, we can process testing images in any arbitrary size. However, as we trained our models on a specific range of effective receptive fields (i.e., our input image size is 256), processing images with very high resolution may cause artifacts. To that end, we follow the post-processing approach used in \cite{afifi2020learning} to deal with high-resolution images (e.g., 16-megapixel) without affecting the quality of the recolored image.

\begin{figure*}
\includegraphics[width=\linewidth]{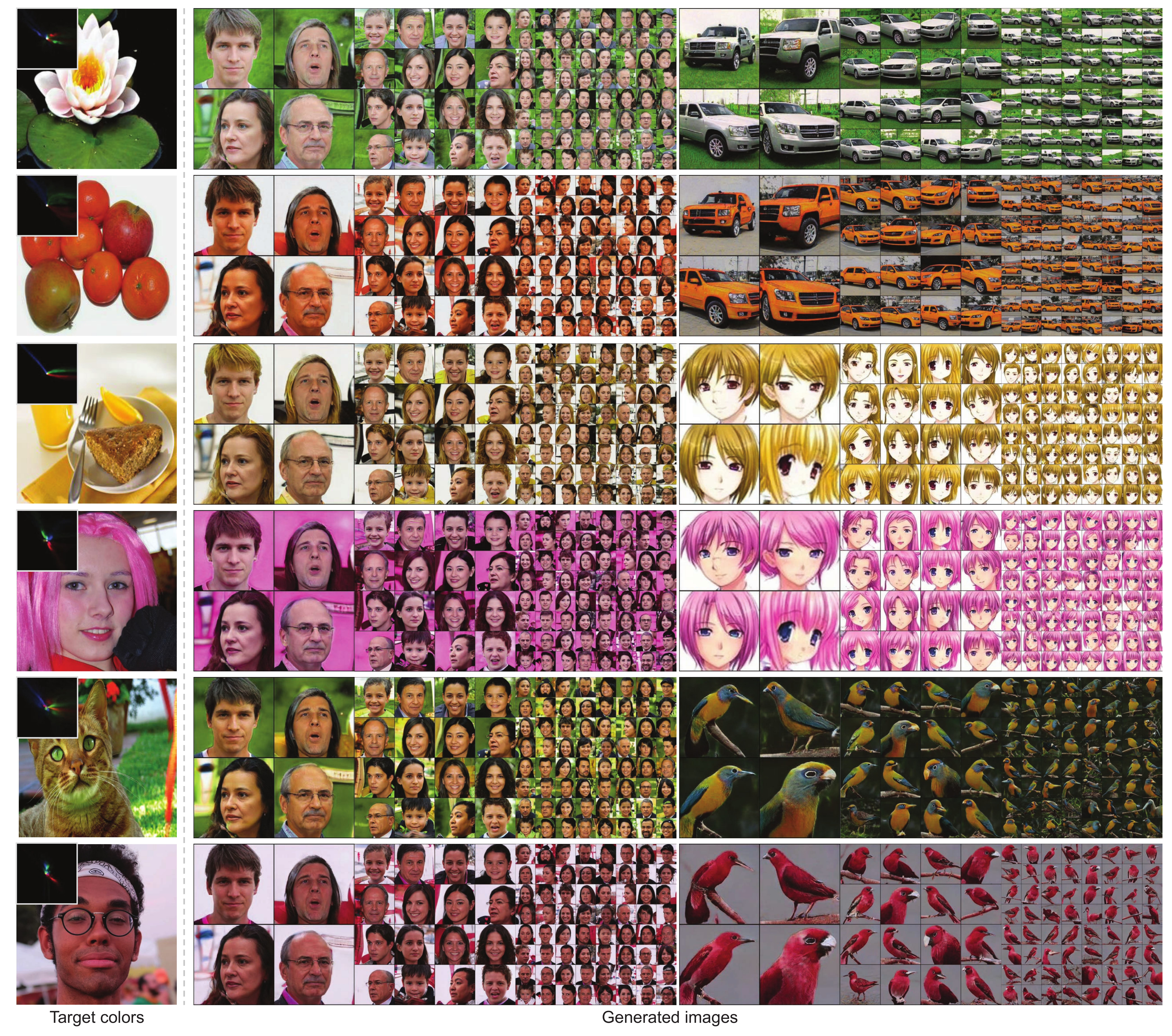}
\vspace{-5.5mm}
\caption{Additional examples of generated images using our HistoGAN. The target colors histograms were computed from the shown target images (left column).}\vspace{-2mm}
\label{fig:qualitative_GAN_results_supp}
\end{figure*}

Specifically, we resize the input image to $256\!\times\!256$ pixels before processing it with our network. Afterward, we apply the bilateral guided upsampling \cite{chen2016bilateral} to construct the mapping from the resized input image and our recoloring result. Then, we apply the constructed bilateral grid to the input image in its original dimensions. Figure \ref{fig:dealing_with_any_size} shows an example of our recoloring result for a high-resolution image ($2048\!\times\!2048$ pixels). As can be seen, our result has the same resolution as the input image with no artifacts.

\begin{figure*}
\includegraphics[width=\linewidth]{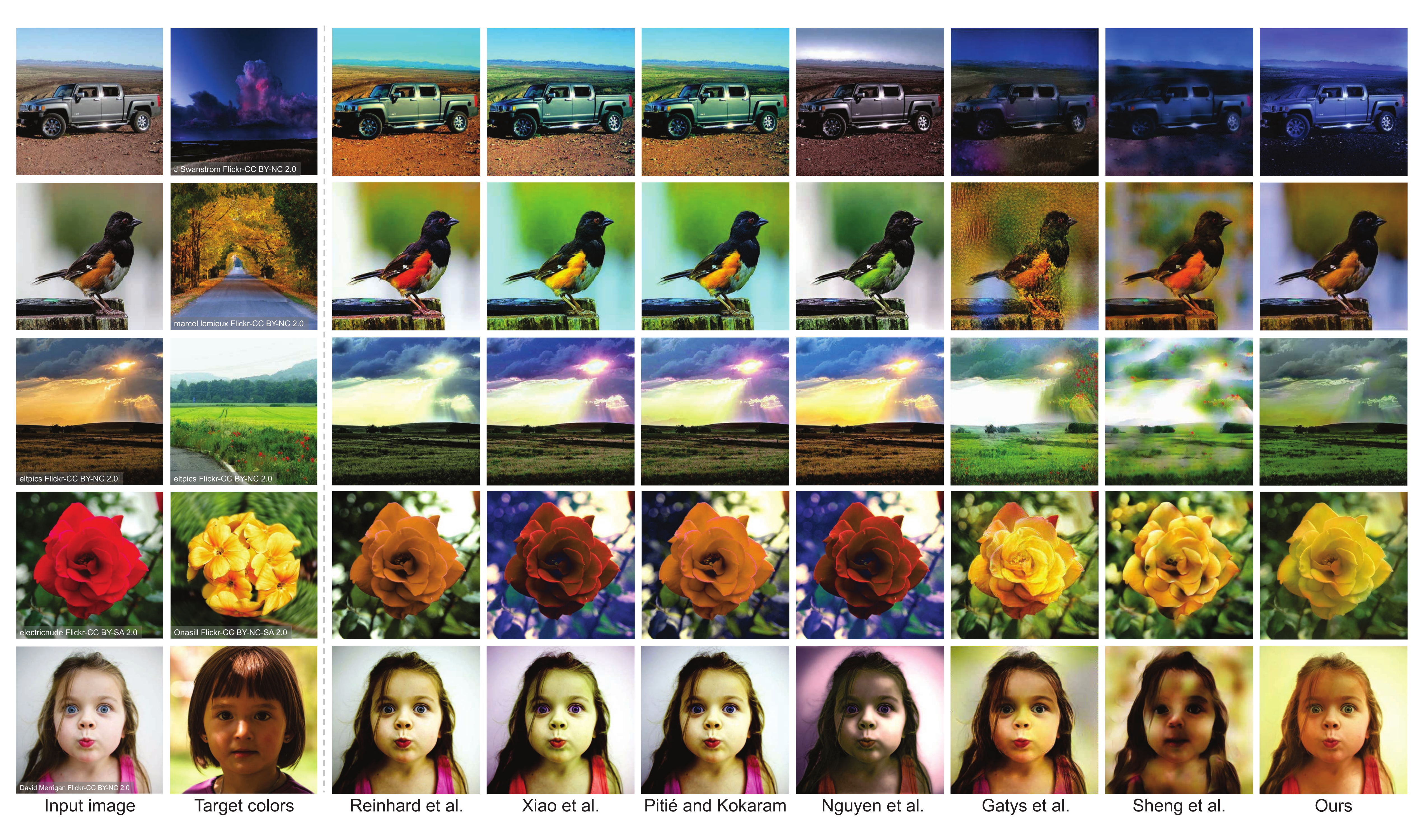}
\vspace{-5.5mm}
\caption{Additional comparisons with image recoloring/style transfer methods. We compare our results with results of the following methods: Reinhard et al., \cite{reinhard2001color}, Xiao et al., \cite{xiao2006color}, Piti\'e and Kokaram \cite{pitie2007}, Nguyen et al., \cite{nguyen2014illuminant}, and Sheng et al., \cite{sheng2018avatar}.}\vspace{-2mm}
\label{fig:compariosns_recoloring_class_based_supp}
\end{figure*}

\begin{figure*}
\includegraphics[width=\linewidth]{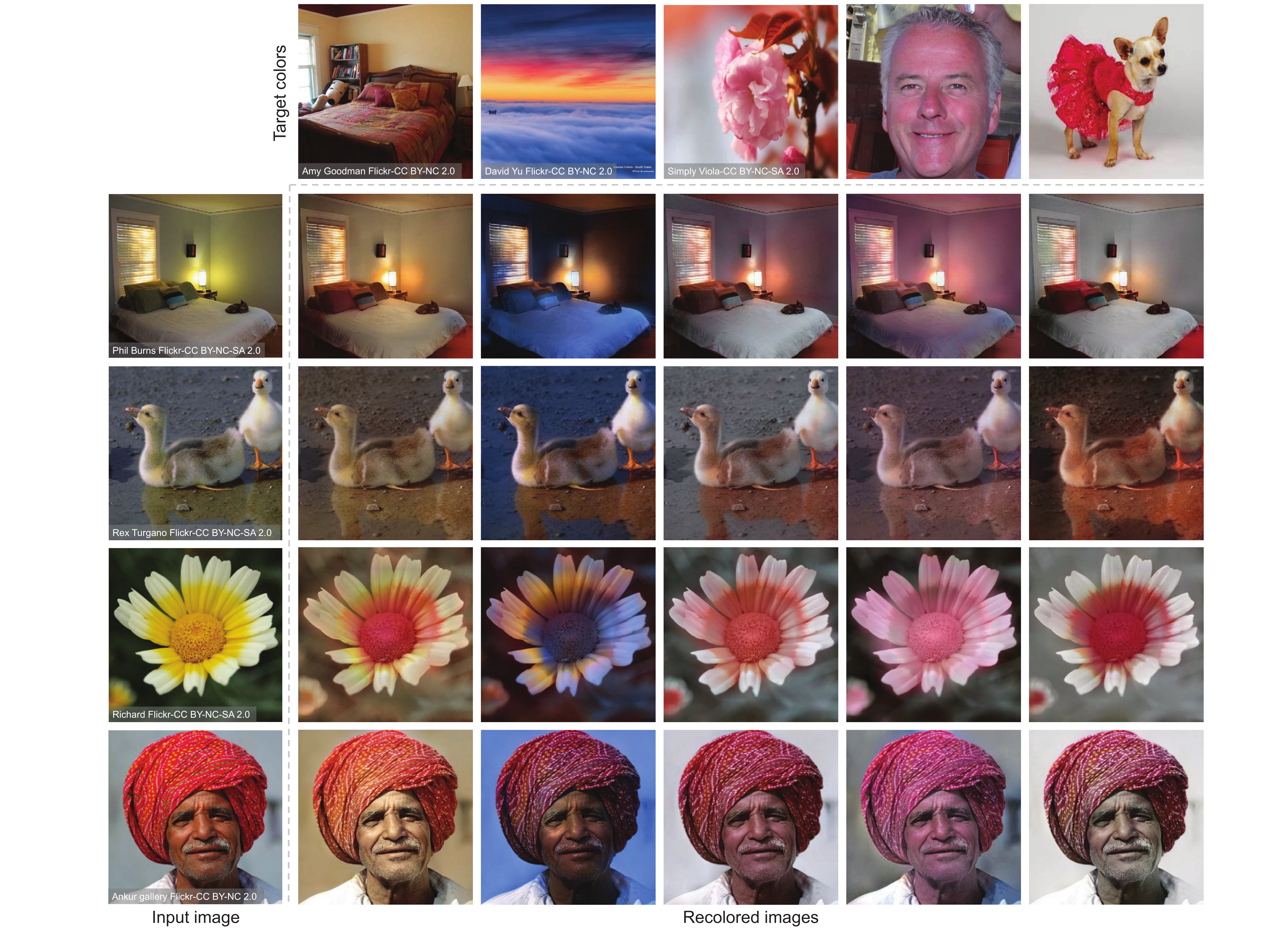}
\vspace{-5.5mm}
\caption{Additional results for image recoloring. We recolor input images, shown in the right by feeding our network with the target histograms of images shown in the top.}\vspace{-1mm}
\label{fig:qualitative_recoloring_class_based}
\end{figure*}

\subsection{Additional Results} \label{sec:additional-results}

This section provides additional results generated by our HistoGAN and ReHistoGAN. As discussed in the main paper, we trained a regression ResNet \cite{he2016deep} model to learn the back-projection from the generated images into the corresponding fine-style vectors of StyleGAN \cite{karras2020analyzing}. This regression model was used to compareHistoGAN and StyleGAN's ability to control the generated images' colors given a target color distribution. Figure \ref{fig:compariosn_w_styleGAN} shows a qualitative comparison between the results of our HistoGAN and StyleGAN with this projection approach. We show additional qualitative comparisons with the recent MixNMatch method \cite{li2020mixnmatch} in Fig.\ \ref{fig:comparison_w_MixNMatch_supp}. In the shown figures, we show the KL divergence and the Hellinger distance between the histograms of the GAN-generated images and the target histogram.

Our HistoGAN, along with the sampling procedure (used for auto recoloring in the main paper) can be used to turn our HistoGAN into a traditional GAN method, where there is no need for any user intervention to input the target histograms. Figure \ref{fig:histGAN_as_styleGAN_supp} shows an example of using our sampling procedure to generate random histogram samples. The generated histogram samples are used by HistoGAN to generate ``unlimited'' number of samples. In the shown figure, we compare between our HistoGAN results, using the generated histograms, with StyleGAN \cite{karras2020analyzing}. In Fig.\ref{fig:histGAN_as_styleGAN_supp}-(A), each row shows generated examples with a fixed fine-style vectors, which are used by the last two blocks of the StyleGAN as these blocks are shown to control the fine-style (e.g., colors, lighting, etc.) of the generated image \cite{karras2019style, karras2020analyzing}.  In Fig.\ref{fig:histGAN_as_styleGAN_supp}-(B), each row shows generated images using our HistoGAN with a fixed generated histogram. As shown in the figure, our HistoGAN generates samples with a higher color diversity compared to StyleGAN results.

Figure \ref{fig:qualitative_GAN_results_supp} shows additional HistoGAN-generated images from different domains. In each row, we show example images generated using the corresponding input target colors. We fixed the coarse- and middle-style vectors, for each domain, to show the response of our HistoGAN to changes in the target histograms.

\begin{figure*}
\includegraphics[width=\linewidth]{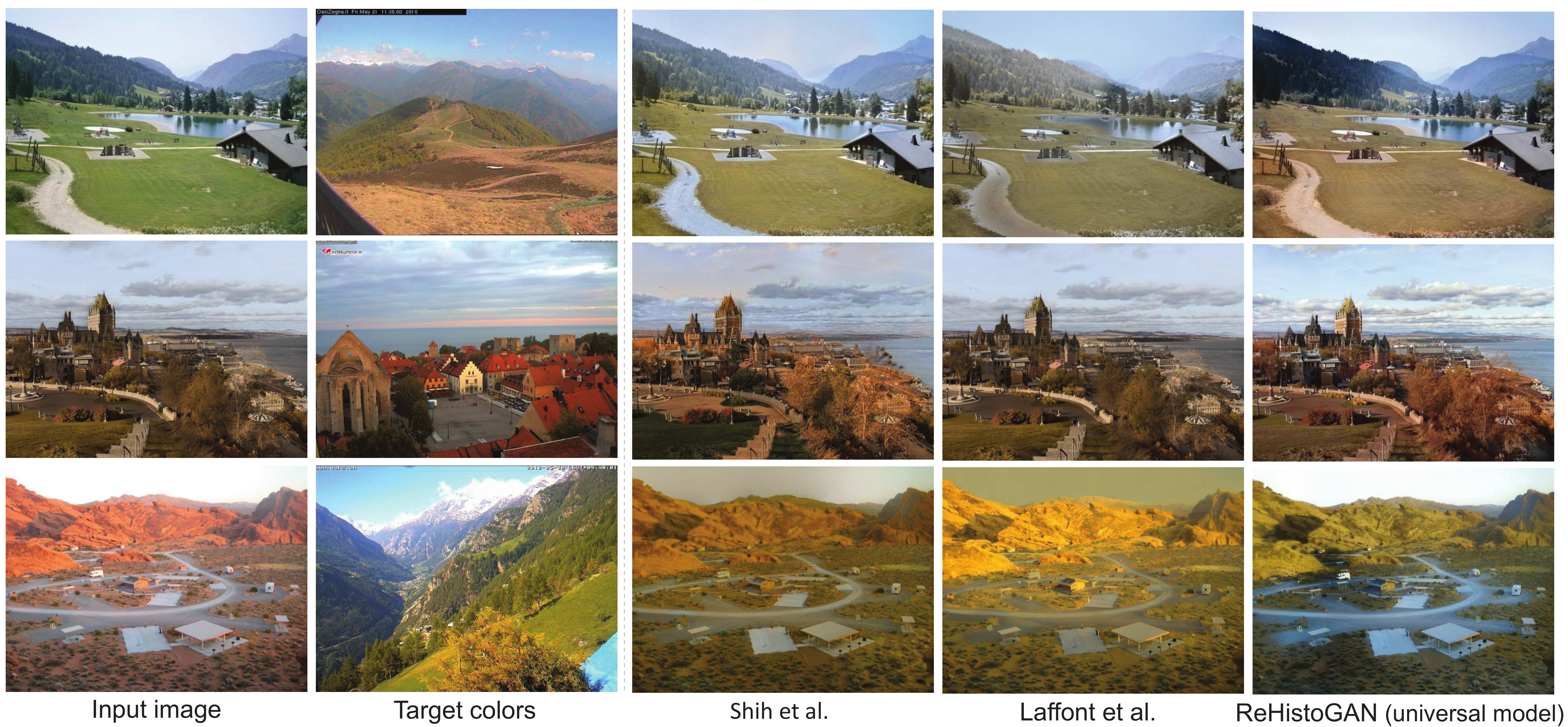}
\vspace{-5.5mm}
\caption{Comparisons between our universal ReHistoGAN and the methods proposed by Shih et al., \cite{shih2013data} and Laffont et al., \cite{laffont2014transient} for color transfer.}\vspace{-2mm}
\label{fig:comparions_universal_model}
\end{figure*}

\begin{figure*}
\includegraphics[width=\linewidth]{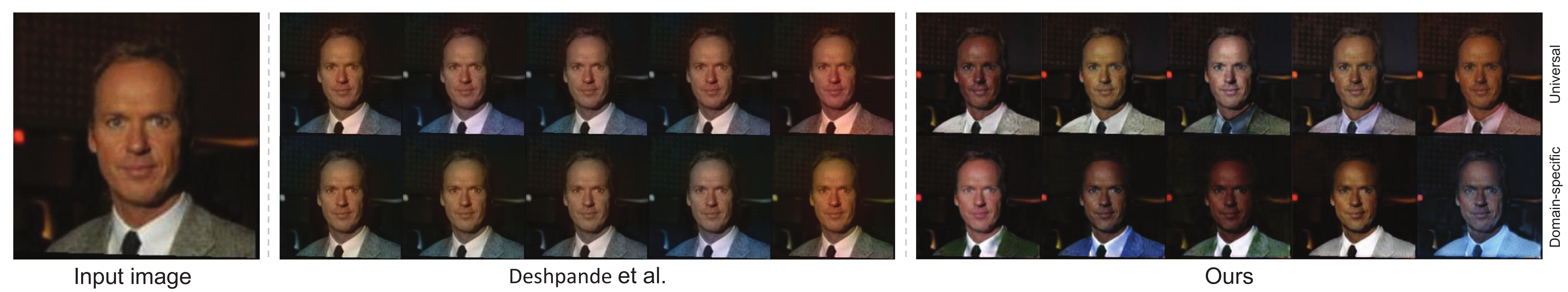}
\vspace{-5.5mm}
\caption{Comparisons between our ReHistoGAN and the diverse colorization method proposed by Deshpande et al., \cite{deshpande2017learning}. For our ReHistoGAN, we show the resutls of our domain-specific and universal models.}\vspace{-2mm}
\label{fig:comparison_with_diverse_colorization}
\end{figure*}

\begin{figure*}
\includegraphics[width=\linewidth]{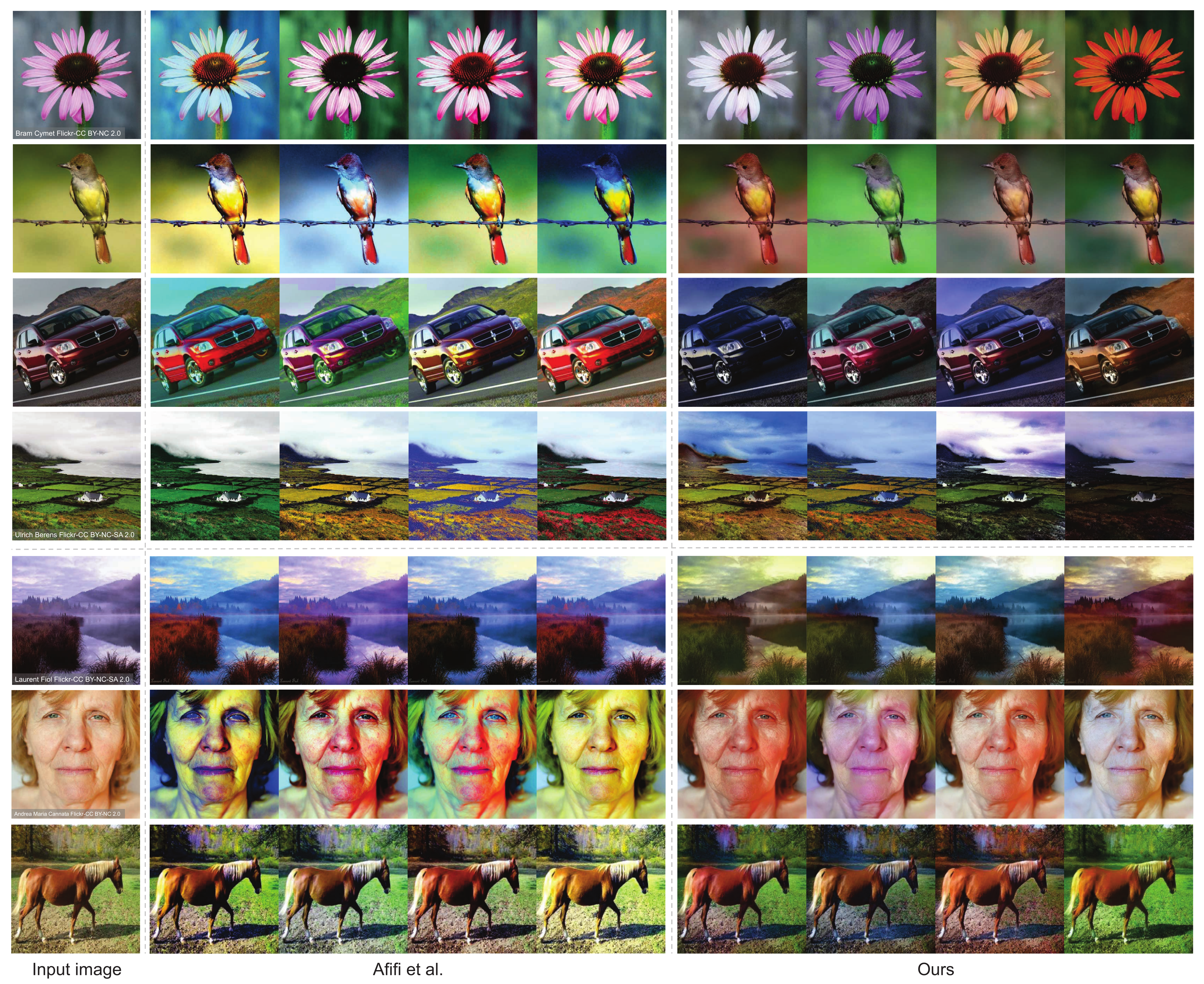}
\vspace{-5.5mm}
\caption{Comparison with the auto image recoloring method proposed in \cite{afifi2019image}. Our recoloring results were produced by domain-specific networks, except for the last three rows, where our results were generated by the universal model.}\vspace{-2mm}
\label{fig:compariosns_auto_recoloring_supp}
\end{figure*}

In the main paper, we showed comparisons with different image recoloring and style transfer methods. Figure \ref{fig:compariosns_recoloring_class_based_supp} shows additional qualitative comparisons. Note that Gaty et al.'s optimization method \cite{gatys2015neural} takes $\sim$4 minutes to process a single image. In contrast, our ReHistoGAN processes a single image in $\sim$0.5 seconds without the guided upsampling procedure \cite{chen2016bilateral}, and $\sim$22 seconds with an unoptimized implementation of the guided upsampling using a single GTX 1080 GPU. Further qualitative examples are shown in Fig.\ \ref{fig:qualitative_recoloring_class_based}. As can be seen, our ReHistoGAN successfully transfers the target colors to the recolored images naturally.

As mentioned in the main paper, there are a few attempts to achieve auto recoloring (e.g., \cite{laffont2014transient, deshpande2017learning, afifi2019image, anokhin2020high}). The high-resolution daytime translation (HiDT) method \cite{anokhin2020high}, for example, achieves the auto-style transfer by sampling from a pre-defined set of target styles. We compared our method and the HiDT method in the main paper, where we used one of the pre-defined target styles as our target histogram. This idea of having a pre-defined set of target styles was originally proposed in \cite{laffont2014transient}, where a set of transient attributes are used to search in a dataset of different target styles. These methods, however, are restricted to the semantic content of the target styles to match the semantic content of training/testing input images. Unlike these auto recoloring/style transfer methods, our ReHistoGAN can deal with histograms taken from any arbitrary domain, as shown in our results in the main paper and these supplementary materials. In Fig.\ \ref{fig:comparions_universal_model}, we show qualitative comparisons of the recoloring results using our universal ReHistoGAN and the method proposed in \cite{laffont2014transient}.

Another strategy for image recoloring is to learn a diverse colorization model. That is, the input image is converted to grayscale, and then a trained method for diverse colorization can generate different colorized versions of the input image. In Fig.\ \ref{fig:comparison_with_diverse_colorization}, we show a qualitative comparison with the diverse colorization method proposed by Deshpande et al., \cite{deshpande2017learning}.

Lastly, we show additional qualitative comparisons with the recent auto-recoloring method proposed by Afifi et al., \cite{afifi2019image} in Fig.\ \ref{fig:compariosns_auto_recoloring_supp}. The figure shows the results of domain-specific ReHistoGAN models (the first four rows) and the universal ReHistoGAN model (the last three rows). As can be seen from the shown figures, our ReHistoGAN arguably produces more realistic recoloring compared to the recoloring results produced by other auto recoloring methods.


{\small
\begin{thebibliography}{10}\itemsep=-1pt

\bibitem{abdal2019image2stylegan}
Rameen Abdal, Yipeng Qin, and Peter Wonka.
\newblock Image2{S}tyle{G}{A}{N}: How to embed images into the stylegan latent
  space?
\newblock In {\em ICCV}, 2019.

\bibitem{abdal2020styleflow}
Rameen Abdal, Peihao Zhu, Niloy Mitra, and Peter Wonka.
\newblock Style{F}low: Attribute-conditioned exploration of
  {S}tyle{G}{A}{N}-generated images using conditional continuous normalizing
  flows.
\newblock {\em arXiv preprint arXiv:2008.02401}, 2020.

\bibitem{afifi201911k}
Mahmoud Afifi.
\newblock 11{K} hands: {G}ender recognition and biometric identification using
  a large dataset of hand images.
\newblock {\em Multimedia Tools and Applications}, 78(15):20835--20854, 2019.

\bibitem{afifi2019sensor}
Mahmoud Afifi and Michael~S Brown.
\newblock Sensor-independent illumination estimation for dnn models.
\newblock In {\em BMVC}, 2019.

\bibitem{afifi2020learning}
Mahmoud Afifi, Konstantinos~G Derpanis, Bj{\"o}rn Ommer, and Michael~S Brown.
\newblock Learning multi-scale photo exposure correction.
\newblock In {\em CVPR}, 2021.

\bibitem{afifi2019color}
Mahmoud Afifi, Brian Price, Scott Cohen, and Michael~S Brown.
\newblock When color constancy goes wrong: Correcting improperly white-balanced
  images.
\newblock In {\em CVPR}, 2019.

\bibitem{afifi2019image}
Mahmoud Afifi, Brian~L Price, Scott Cohen, and Michael~S Brown.
\newblock Image recoloring based on object color distributions.
\newblock In {\em Eurographics 2019 (short papers)}, 2019.

\bibitem{anokhin2020high}
Ivan Anokhin, Pavel Solovev, Denis Korzhenkov, Alexey Kharlamov, Taras
  Khakhulin, Aleksei Silvestrov, Sergey Nikolenko, Victor Lempitsky, and Gleb
  Sterkin.
\newblock High-resolution daytime translation without domain labels.
\newblock In {\em CVPR}, 2020.

\bibitem{arjovsky2017wasserstein}
Martin Arjovsky, Soumith Chintala, and L{\'e}on Bottou.
\newblock Wasserstein {G}{A}{N}.
\newblock {\em arXiv preprint arXiv:1701.07875}, 2017.

\bibitem{avi2020deephist}
Mor Avi-Aharon, Assaf Arbelle, and Tammy~Riklin Raviv.
\newblock Deephist: Differentiable joint and color histogram layers for
  image-to-image translation.
\newblock {\em arXiv preprint arXiv:2005.03995}, 2020.

\bibitem{CCC}
Jonathan~T Barron.
\newblock Convolutional color constancy.
\newblock In {\em ICCV}, 2015.

\bibitem{fivek}
Vladimir Bychkovsky, Sylvain Paris, Eric Chan, and Fr{\'e}do Durand.
\newblock Learning photographic global tonal adjustment with a database of
  input / output image pairs.
\newblock In {\em CVPR}, 2011.

\bibitem{chang2015palette}
Huiwen Chang, Ohad Fried, Yiming Liu, Stephen DiVerdi, and Adam Finkelstein.
\newblock Palette-based photo recoloring.
\newblock {\em ACM Transactions on Graphics (TOG)}, 34(4):139--1, 2015.

\bibitem{chen2016bilateral}
Jiawen Chen, Andrew Adams, Neal Wadhwa, and Samuel~W Hasinoff.
\newblock Bilateral guided upsampling.
\newblock {\em ACM Transactions on Graphics (TOG)}, 35(6):1--8, 2016.

\bibitem{choi2020stargan}
Yunjey Choi, Youngjung Uh, Jaejun Yoo, and Jung-Woo Ha.
\newblock Star{G}{A}{N} {V}2: Diverse image synthesis for multiple domains.
\newblock In {\em CVPR}, 2020.

\bibitem{animedataset}
Spencer Churchill.
\newblock Anime face dataset.
\newblock \url{ https://www.kaggle.com/splcher/animefacedataset}.
\newblock [Online; accessed October 27, 2020].

\bibitem{catdataset}
Chris Crawford and NAIN.
\newblock Cat dataset.
\newblock \url{https://www.kaggle.com/crawford/cat-dataset}.
\newblock [Online; accessed October 27, 2020].

\bibitem{deshpande2017learning}
Aditya Deshpande, Jiajun Lu, Mao-Chuang Yeh, Min Jin~Chong, and David Forsyth.
\newblock Learning diverse image colorization.
\newblock In {\em CVPR}, 2017.

\bibitem{8939562}
S.~R. {Dubey}, S. {Chakraborty}, S.~K. {Roy}, S. {Mukherjee}, S.~K. {Singh},
  and B.~B. {Chaudhuri}.
\newblock diff{G}rad: An optimization method for convolutional neural networks.
\newblock {\em IEEE Transactions on Neural Networks and Learning Systems},
  31(11):4500--4511, 2020.

\bibitem{eibenberger2012importance}
Eva Eibenberger and Elli Angelopoulou.
\newblock The importance of the normalizing channel in log-chromaticity space.
\newblock In {\em CIP}, 2012.

\bibitem{faridul2016colour}
H~Sheikh Faridul, Tania Pouli, Christel Chamaret, J{\"u}rgen Stauder, Erik
  Reinhard, Dmitry Kuzovkin, and Alain Tr{\'e}meau.
\newblock Colour mapping: A review of recent methods, extensions and
  applications.
\newblock In {\em Computer Graphics Forum}, 2016.

\bibitem{finlayson2001color}
Graham~D Finlayson and Steven~D Hordley.
\newblock Color constancy at a pixel.
\newblock {\em JOSA A}, 2001.

\bibitem{gatys2015neural}
Leon~A Gatys, Alexander~S Ecker, and Matthias Bethge.
\newblock A neural algorithm of artistic style.
\newblock {\em arXiv preprint arXiv:1508.06576}, 2015.

\bibitem{gatys2016image}
Leon~A Gatys, Alexander~S Ecker, and Matthias Bethge.
\newblock Image style transfer using convolutional neural networks.
\newblock In {\em CVPR}, 2016.

\bibitem{goodfellow2014generative}
Ian Goodfellow, Jean Pouget-Abadie, Mehdi Mirza, Bing Xu, David Warde-Farley,
  Sherjil Ozair, Aaron Courville, and Yoshua Bengio.
\newblock Generative adversarial nets.
\newblock In {\em NeurIPS}, 2014.

\bibitem{he2015delving}
Kaiming He, Xiangyu Zhang, Shaoqing Ren, and Jian Sun.
\newblock Delving deep into rectifiers: Surpassing human-level performance on
  {I}mage{N}et classification.
\newblock In {\em ICCV}, 2015.

\bibitem{he2016deep}
Kaiming He, Xiangyu Zhang, Shaoqing Ren, and Jian Sun.
\newblock Deep residual learning for image recognition.
\newblock In {\em CVPR}, 2016.

\bibitem{he2019progressive}
Mingming He, Jing Liao, Dongdong Chen, Lu Yuan, and Pedro~V Sander.
\newblock Progressive color transfer with dense semantic correspondences.
\newblock {\em ACM Transactions on Graphics (TOG)}, 38(2):1--18, 2019.

\bibitem{heusel2017gans}
Martin Heusel, Hubert Ramsauer, Thomas Unterthiner, Bernhard Nessler, and Sepp
  Hochreiter.
\newblock {G}{A}{N}s trained by a two time-scale update rule converge to a
  local nash equilibrium.
\newblock In {\em NeurIPS}, 2017.

\bibitem{isola2017image}
Phillip Isola, Jun-Yan Zhu, Tinghui Zhou, and Alexei~A Efros.
\newblock Image-to-image translation with conditional adversarial networks.
\newblock In {\em CVPR}, 2017.

\bibitem{johnson2016perceptual}
Justin Johnson, Alexandre Alahi, and Li Fei-Fei.
\newblock Perceptual losses for real-time style transfer and super-resolution.
\newblock In {\em ECCV}, 2016.

\bibitem{karras2017progressive}
Tero Karras, Timo Aila, Samuli Laine, and Jaakko Lehtinen.
\newblock Progressive growing of {G}{A}{N}s for improved quality, stability,
  and variation.
\newblock {\em arXiv preprint arXiv:1710.10196}, 2017.

\bibitem{karras2019style}
Tero Karras, Samuli Laine, and Timo Aila.
\newblock A style-based generator architecture for generative adversarial
  networks.
\newblock In {\em CVPR}, 2019.

\bibitem{karras2020analyzing}
Tero Karras, Samuli Laine, Miika Aittala, Janne Hellsten, Jaakko Lehtinen, and
  Timo Aila.
\newblock Analyzing and improving the image quality of {S}tyle{G}{A}{N}.
\newblock In {\em CVPR}, 2020.

\bibitem{khosla2011novel}
Aditya Khosla, Nityananda Jayadevaprakash, Bangpeng Yao, and Fei-Fei Li.
\newblock Novel dataset for fine-grained image categorization: {S}tanford dogs.
\newblock In {\em CVPR Workshops}, 2011.

\bibitem{krause20133d}
Jonathan Krause, Michael Stark, Jia Deng, and Li Fei-Fei.
\newblock 3{D} object representations for fine-grained categorization.
\newblock In {\em ICCV Workshops}, 2013.

\bibitem{kuznetsova2020open}
Alina Kuznetsova, Hassan Rom, Neil Alldrin, Jasper Uijlings, Ivan Krasin, Jordi
  Pont-Tuset, Shahab Kamali, Stefan Popov, Matteo Malloci, Alexander
  Kolesnikov, et~al.
\newblock The open images dataset {V}4.
\newblock {\em International Journal of Computer Vision}, pages 1--26, 2020.

\bibitem{laffont2014transient}
Pierre-Yves Laffont, Zhile Ren, Xiaofeng Tao, Chao Qian, and James Hays.
\newblock Transient attributes for high-level understanding and editing of
  outdoor scenes.
\newblock {\em ACM Transactions on graphics (TOG)}, 33(4):1--11, 2014.

\bibitem{li2020mixnmatch}
Yuheng Li, Krishna~Kumar Singh, Utkarsh Ojha, and Yong~Jae Lee.
\newblock Mix{N}{M}atch: {M}ultifactor disentanglement and encoding for
  conditional image generation.
\newblock In {\em CVPR}, 2020.

\bibitem{lin2014microsoft}
Tsung-Yi Lin, Michael Maire, Serge Belongie, James Hays, Pietro Perona, Deva
  Ramanan, Piotr Doll{\'a}r, and C~Lawrence Zitnick.
\newblock Microsoft {C}{O}{C}{O}: Common objects in context.
\newblock In {\em ECCV}, 2014.

\bibitem{liu2019wasserstein}
Huidong Liu, Xianfeng Gu, and Dimitris Samaras.
\newblock Wasserstein {G}{A}{N} with quadratic transport cost.
\newblock In {\em ICCV}, 2019.

\bibitem{liu2015deep}
Ziwei Liu, Ping Luo, Xiaogang Wang, and Xiaoou Tang.
\newblock Deep learning face attributes in the wild.
\newblock In {\em ICCV}, 2015.

\bibitem{luan2017deep}
Fujun Luan, Sylvain Paris, Eli Shechtman, and Kavita Bala.
\newblock Deep photo style transfer.
\newblock In {\em CVPR}, 2017.

\bibitem{maas2013rectifier}
Andrew~L Maas, Awni~Y Hannun, and Andrew~Y Ng.
\newblock Rectifier nonlinearities improve neural network acoustic models.
\newblock In {\em ICML}, 2013.

\bibitem{maggiori2017can}
Emmanuel Maggiori, Yuliya Tarabalka, Guillaume Charpiat, and Pierre Alliez.
\newblock Can semantic labeling methods generalize to any city? {T}he inria
  aerial image labeling benchmark.
\newblock In {\em International Geoscience and Remote Sensing Symposium
  (IGARSS)}, 2017.

\bibitem{nguyen2014illuminant}
Rang~MH Nguyen, Seon~Joo Kim, and Michael~S Brown.
\newblock Illuminant aware gamut-based color transfer.
\newblock In {\em Computer Graphics Forum}, 2014.

\bibitem{nilsback2008automated}
Maria-Elena Nilsback and Andrew Zisserman.
\newblock Automated flower classification over a large number of classes.
\newblock In {\em Indian Conference on Computer Vision, Graphics \& Image
  Processing}, 2008.

\bibitem{perez2003poisson}
Patrick P{\'e}rez, Michel Gangnet, and Andrew Blake.
\newblock Poisson image editing.
\newblock In {\em SIGGRAPH}. 2003.

\bibitem{pidhorskyi2020adversarial}
Stanislav Pidhorskyi, Donald~A Adjeroh, and Gianfranco Doretto.
\newblock Adversarial latent autoencoders.
\newblock In {\em CVPR}, 2020.

\bibitem{pitie2007}
F. Pitie and A. Kokaram.
\newblock The linear {M}onge-{K}antorovitch linear colour mapping for
  example-based colour transfer.
\newblock In {\em European Conference on Visual Media Production}, 2007.

\bibitem{radford2015unsupervised}
Alec Radford, Luke Metz, and Soumith Chintala.
\newblock Unsupervised representation learning with deep convolutional
  generative adversarial networks.
\newblock {\em arXiv preprint arXiv:1511.06434}, 2015.

\bibitem{reinhard2001color}
Erik Reinhard, Michael Adhikhmin, Bruce Gooch, and Peter Shirley.
\newblock Color transfer between images.
\newblock {\em IEEE Computer graphics and applications}, 21(5):34--41, 2001.

\bibitem{ronneberger2015u}
Olaf Ronneberger, Philipp Fischer, and Thomas Brox.
\newblock U-{N}et: Convolutional networks for biomedical image segmentation.
\newblock In {\em MICCAI}, 2015.

\bibitem{saito2020coco}
Kuniaki Saito, Kate Saenko, and Ming-Yu Liu.
\newblock {C}{O}{C}{O}-{F}{U}{N}{I}{T}: Few-shot unsupervised image translation
  with a content conditioned style encoder.
\newblock In {\em ECCV}, 2020.

\bibitem{shaham2019singan}
Tamar~Rott Shaham, Tali Dekel, and Tomer Michaeli.
\newblock Sin{G}{A}{N}: Learning a generative model from a single natural
  image.
\newblock In {\em ICCV}, 2019.

\bibitem{sheng2018avatar}
Lu Sheng, Ziyi Lin, Jing Shao, and Xiaogang Wang.
\newblock Avatar-{N}et: Multi-scale zero-shot style transfer by feature
  decoration.
\newblock In {\em CVPR}, 2018.

\bibitem{shih2013data}
Yichang Shih, Sylvain Paris, Fr{\'e}do Durand, and William~T Freeman.
\newblock Data-driven hallucination of different times of day from a single
  outdoor photo.
\newblock {\em ACM Transactions on Graphics (TOG)}, 32(6):1--11, 2013.

\bibitem{shocher2020semantic}
Assaf Shocher, Yossi Gandelsman, Inbar Mosseri, Michal Yarom, Michal Irani,
  William~T Freeman, and Tali Dekel.
\newblock Semantic pyramid for image generation.
\newblock In {\em CVPR}, 2020.

\bibitem{shugrina2020nonlinear}
Maria Shugrina, Amlan Kar, Sanja Fidler, and Karan Singh.
\newblock Nonlinear color triads for approximation, learning and direct
  manipulation of color distributions.
\newblock {\em ACM Transactions on Graphics (TOG)}, 39(4):97--1, 2020.

\bibitem{szegedy2015going}
Christian Szegedy, Wei Liu, Yangqing Jia, Pierre Sermanet, Scott Reed, Dragomir
  Anguelov, Dumitru Erhan, Vincent Vanhoucke, and Andrew Rabinovich.
\newblock Going deeper with convolutions.
\newblock In {\em CVPR}, 2015.

\bibitem{ulyanov2016instance}
Dmitry Ulyanov, Andrea Vedaldi, and Victor Lempitsky.
\newblock Instance normalization: The missing ingredient for fast stylization.
\newblock {\em arXiv preprint arXiv:1607.08022}, 2016.

\bibitem{wah2011caltech}
Catherine Wah, Steve Branson, Peter Welinder, Pietro Perona, and Serge
  Belongie.
\newblock The {C}altech-{U}{C}{S}{D} birds-200-2011 dataset.
\newblock 2011.

\bibitem{xiao2006color}
Xuezhong Xiao and Lizhuang Ma.
\newblock Color transfer in correlated color space.
\newblock In {\em International conference on Virtual reality continuum and its
  applications}, 2006.

\bibitem{yu2015lsun}
Fisher Yu, Ari Seff, Yinda Zhang, Shuran Song, Thomas Funkhouser, and Jianxiong
  Xiao.
\newblock {L}{S}{U}{N}: {C}onstruction of a large-scale image dataset using
  deep learning with humans in the loop.
\newblock {\em arXiv preprint arXiv:1506.03365}, 2015.

\bibitem{zhang2017palette}
Qing Zhang, Chunxia Xiao, Hanqiu Sun, and Feng Tang.
\newblock Palette-based image recoloring using color decomposition
  optimization.
\newblock {\em IEEE Transactions on Image Processing}, 26(4):1952--1964, 2017.

\bibitem{zhou2017scene}
Bolei Zhou, Hang Zhao, Xavier Puig, Sanja Fidler, Adela Barriuso, and Antonio
  Torralba.
\newblock Scene parsing through {A}{D}{E}20{K} dataset.
\newblock In {\em CVPR}, 2017.

\bibitem{zhou2019semantic}
Bolei Zhou, Hang Zhao, Xavier Puig, Tete Xiao, Sanja Fidler, Adela Barriuso,
  and Antonio Torralba.
\newblock Semantic understanding of scenes through the {A}{D}{E}20{K} dataset.
\newblock {\em International Journal of Computer Vision}, 127(3):302--321,
  2019.

\end{thebibliography}

}
\end{document}